\documentclass[a4paper,fleqn]{cas-dc}

\usepackage{ulem}

\usepackage[numbers]{natbib}
\usepackage{makecell}
\usepackage{pifont}
\def\tsc#1{\csdef{#1}{\textsc{\lowercase{#1}}\xspace}}
\tsc{WGM}
\tsc{QE}
\tsc{EP}
\tsc{PMS}
\tsc{BEC}
\tsc{DE}

\begin{document}
\let\WriteBookmarks\relax
\def\floatpagepagefraction{1}
\def\textpagefraction{.001}
\shorttitle{DDFP: Data-dependent Frequency Prompt for Source Free Domain Adaptation of Medical Image Segmentation}
\shortauthors{S. Yin, S. Liu, M. Wang}

\title [mode = title]{DDFP: Data-dependent Frequency Prompt for Source Free Domain Adaptation of Medical Image Segmentation}                      
\tnotemark[1,2]



\author[1, 2]{Siqi Yin}
\ead{sqyin21@m.fudan.edu.cn}


\affiliation[1]{organization={Digital Medical Research Center, School of Basic Medical 
Science, Fudan University},
                city={Shanghai},
                postcode={200032}, 
                country={China}}

\affiliation[2]{organization={Shanghai Key Laboratory of Medical Image Computing and Computer Assisted Intervention},
                city={Shanghai},
                postcode={200032}, 
                country={China}}

\author[1, 2]{Shaolei Liu}
\ead{slliu@fudan.edu.cn}

\author[1, 2]{Manning Wang}
\cormark[1]
\ead{mnwang@fudan.edu.cn}

\cortext[cor1]{Corresponding author}



  \begin{abstract}
    Domain adaptation addresses the challenge of model performance degradation caused by domain gaps. In the typical setup for unsupervised domain adaptation, labeled data from a source domain and unlabeled data from a target domain are used to train a target model. However, access to labeled source domain data, particularly in medical datasets, can be restricted due to privacy policies. As a result, research has increasingly shifted to source-free domain adaptation (SFDA), which requires only a pretrained model from the source domain and unlabeled data from the target domain data for adaptation. Existing SFDA methods often rely on domain-specific image style translation and self-supervision techniques to bridge the domain gap and train the target domain model. However, the quality of domain-specific style-translated images and pseudo-labels produced by these methods still leaves room for improvement. Moreover, training the entire model during adaptation can be inefficient under limited supervision. In this paper, we propose a novel SFDA framework to address these challenges. Specifically, to effectively mitigate the impact of domain gap in the initial training phase, we introduce preadaptation to generate a preadapted model, which serves as an initialization of target model and allows for the generation of high-quality enhanced pseudo-labels without introducing extra parameters. Additionally, we propose a data-dependent frequency prompt to more effectively translate target domain images into a source-like style. To further enhance adaptation, we employ a style-related layer fine-tuning strategy, specifically designed for SFDA, to train the target model using the prompted target domain images and pseudo-labels. Extensive experiments on cross-modality abdominal and cardiac SFDA segmentation tasks demonstrate that our proposed method outperforms existing state-of-the-art methods. Our code is available online.
  \end{abstract}



\begin{keywords}
  Semantic segmentation \sep Domain adaptation \sep Medical image \sep Prompt learning 
\end{keywords}

\maketitle

\section{Introduction}

Deep learning has become widely used in the field of medical image analysis, and its promising performance largely relies on the availability of sufficient labeled data for model training. However, data collection and labeling are labor-intensive and time-consuming, especially for medical image segmentation tasks that require expert annotators for dense annotation. A common solution is to train a model using labeled data from a source domain and then transfer the learned knowledge to a new dataset (target domain), which is often unlabeled \cite{1csurka2017domain}. However, data distributions can differ significantly between domains due to factors such as acquisition protocols and data modalities, creating a domain gap. When a model trained on the source domain is directly applied to the target domain, this gap commonly leads to severe performance degradation \cite{2datashift}. To address this, domain adaptation (DA) has been proposed to improve model performance under domain shifts. As one of the most commonly used settings, unsupervised domain adaptation (UDA) has been extensively studied and shown success in medical tasks such as object detection \cite{3xing2020bidirectional, 43}, classification \cite{41, 45, 46} and segmentation\cite{4sifachen2019synergistic,5hsu2021darcnn,6chen2020unsupervised,7liu2023structure,38_yang2020fda,39_huang2021fsdr,40_liu2023reducing,42,44}. These UDA methods generally require labeled data from the source domain and unlabeled data from the target domain data for target model training. However, this approach becomes impractical when access to source domain is restricted, for example, due to privacy concerns in medical datasets \cite{8yang2021federated}. 

\begin{figure}[!t]
  \centering
  \includegraphics[scale=0.6]{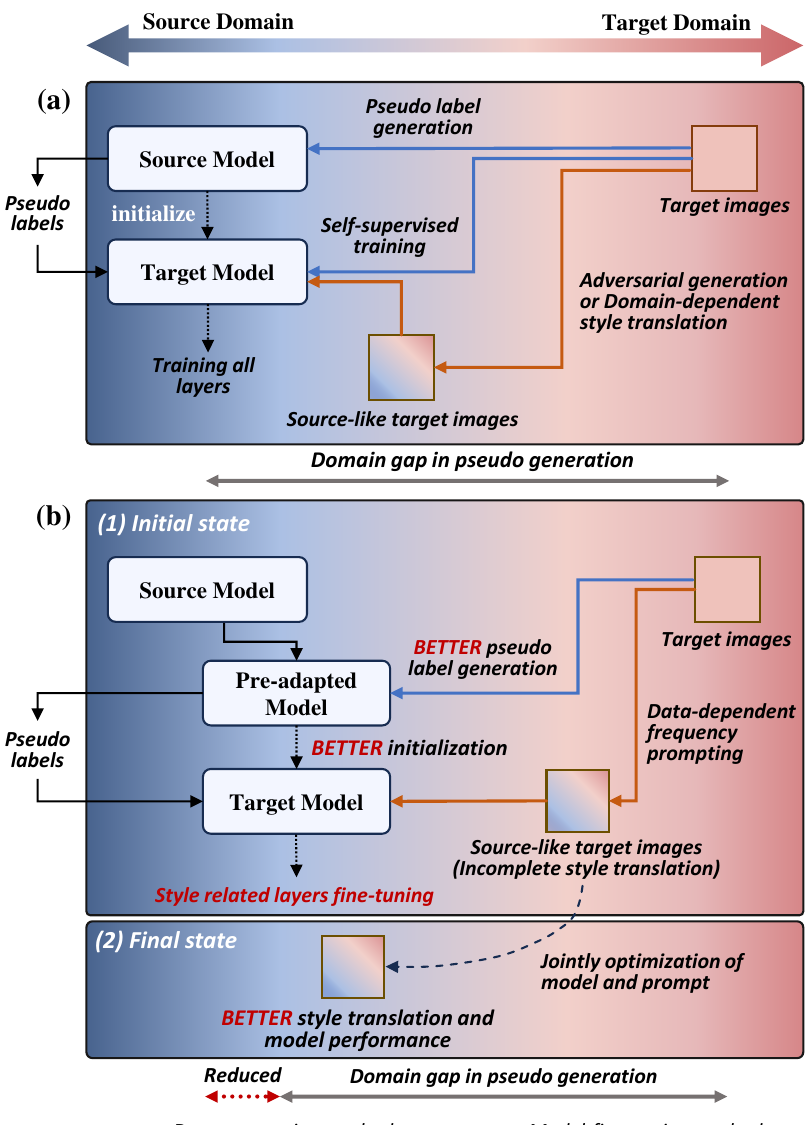}
  \caption{Comparison between (a) the previous source-free domain adaptation framework and (b) the proposed DDFP framework. The DDFP framework reduces the domain gap throughout both the initial and subsequent training phases by utilizing a preadapted model, data-dependent frequency prompt learning, and pseudo-labeling strategies.}
  \label{fig1}
\end{figure}

To alleviate the dependence on source domain data during the adaptation process, the source-free DA (SFDA) scheme is proposed. In this approach, only unlabeled target domain data and a pretrained source domain model (referred to as the ``source model'') are used to train the target domain model (referred to as the ``target model''). In semantic segmentation tasks, SFDA methods typically rely on two strategies:  data generation \cite{9hu2022prosfda, 10yang2022FSM, 11wang2023fvp, 12kbshong2022source, 13liu2021source} and model fine-tuning \cite{9hu2022prosfda, 10yang2022FSM, 12kbshong2022source, 14yu2023source, 15chen2021source, 16bateson2022source}, as shown in Fig. \ref{fig1}(a). The data-generation strategy translates target domain images to a source-like style, reducing the domain gap and either directly improving the source model's performance on the target data \cite{11wang2023fvp} or assisting the training of the target model \cite{10yang2022FSM,9hu2022prosfda,12kbshong2022source,13liu2021source}. Recently, prompt learning has been introduced in SFDA for image style translation, applying a trainable prompt in either the frequency \cite{11wang2023fvp} or spatial domain \cite{9hu2022prosfda} to align target domain images with the source domain style. Meanwhile, the model fine-tuning strategy initializes the target model using the pretrained source model and fine-tunes it using self-supervised techniques, such as pseudo-labeling \cite{10yang2022FSM, 9hu2022prosfda, 15chen2021source}.

Despite promising results from existing SFDA methods, there are three key challenges that hinder further progress. 
\textbf{Problem 1.} Current prompt learning-based style translation methods apply a same prompt across the entire target domain \cite{11wang2023fvp, 9hu2022prosfda}, ignoring intradomain variations. 
\textbf{Problem 2.} In the early stages of training, the domain gap between the target domain data and the source model remains large. Directly using the source model to initialize the target model and generate pseudo-labels can lead to mismatches between the data and the model, negatively impacting both model performance and pseudo-label quality.
\textbf{Problem 3.} SFDA methods that fine-tune the entire target model \cite{9hu2022prosfda, 14yu2023source, 15chen2021source, 10yang2022FSM} or layers before the final classifier \cite{14yu2023source} often face inefficiencies. Research has shown that features from shallow and deep layers correspond to style and content information, respectively \cite{17pan2018two, 10yang2022FSM}, with the domain gap primarily involving low-level stylistic differences \cite{18dou2019pnp}. Fine-tuning the entire model may not be necessary, especially when supervision is limited in SFDA scenarios.

To address the aforementioned challenges, we propose a novel framework utilizing data-dependent frequency prompt (DDFP), as illustrated in Fig. \ref{fig1}(b). Our approach tackles the domain gap at both the initial and subsequence states of training by employing model preadaptation, prompt learning, and pseudo-labeling, thereby improving the efficacy of model transfer across domains. In the initial stage, where the domain gap is large and unaddressed by prior training, we focus on mitigating the gap from the model perspective. Specifically, we calibrate the batch normalization (BN) statistic of the source model to derive a preadapted model that is better aligned with the target domain distribution than the original source model. This strategy offers two key advantages. Initializing the target model with the preadapted model reduces the domain gap between the model and target domain data during the early training stages. Besides, using the preadapted model to generate pseudo-labels for the target domain images improves pseudo-label quality, thus enhancing the final performance of the target model (addressing Problem 2).

During the target model training, we address the domain gap at the image level by applying data-dependent style transfer, combined with pseudo-labeling to train the target model. We introduce the data-dependent frequency prompt to more precisely and individually translate target domain images into a source-like style (addressing Problem 1). We use the prompted target domain images as input and focus on training the style-related layers of the target model, thereby enhancing DA (addressing Problem 3). By leveraging the preadapted model, we generate higher-quality pseudo-labels and impose constraints at the output layer, improving the effectiveness of self-supervised learning. Experiments conducted on cross-modality abdominal and cardiac segmentation tasks show that our method outperforms existing state-of-the-art techniques, achieving a higher average Dice coefficient.

The main contributions of this work are as follows:

\begin{enumerate}
\item	We propose the use of data-dependent frequency prompt to reduce the domain gap in SFDA, enabling better image style translation and significantly improving target model performance.
\item	By utilizing the preadapted source model for target model initialization and pseudo-label generation, we effectively mitigate the domain gap during the initial training phase and enhance pseudo-label quality, leading to improved self-supervised training outcomes.
\item	We introduce a style-related layer fine-tuning strategy tailored for SFDA, which further enhances target model performance with fewer trainable parameters.
\item	Our method demonstrates superior performance in cross-modality DA, particularly on abdominal and cardiac datasets, achieving higher average Dice coefficients than current state-of-the-art methods.
\end{enumerate}

\section{Related work}
\subsection{Source free domain adaptation}

SFDA aims to adapt a model trained on the source domain to a target domain without requiring access to source domain data or target domain labels. For semantic segmentation, existing SFDA methods typically initialize the target model with the source model and then adapt it using two main strategies \cite{20li2024comprehensive, 21wang2022exploring}, model fine-tuning \cite{9hu2022prosfda, 10yang2022FSM, 12kbshong2022source, 14yu2023source, 15chen2021source, 16bateson2022source,48ttsfuda} and data generation \cite{9hu2022prosfda, 10yang2022FSM, 11wang2023fvp, 12kbshong2022source, 13liu2021source}. In model fine-tuning methods, the source model–initialized target model is trained using target domain images along with pseudo-labels \cite{9hu2022prosfda, 10yang2022FSM, 11wang2023fvp, 15chen2021source,48ttsfuda} or other self-supervision techniques such as entropy minimization \cite{12kbshong2022source, 16bateson2022source} or contrastive learning \cite{9hu2022prosfda, 10yang2022FSM, 14yu2023source}. For example, DPL \cite{15chen2021source} generates pseudo-labels for target domain images based on the source model's predictions, which are then refined using feature-to-prototype distance and uncertainty maps before being used for fine-tuning the target model. 
Conversely, data-generation methods focus on creating an intermediate domain (either target-like or source-like) through domain-specific reconstruction \cite{22li2020MA} or style translation  \cite{9hu2022prosfda, 10yang2022FSM, 11wang2023fvp, 12kbshong2022source, 13liu2021source}, reducing the domain gap at the image level and aiding target model training. For instance, 3C-GAN \cite{22li2020MA} employs a conditional generative adversarial network (GAN) to generate target-style images, collaboratively training a classifier and generator using both original target domain images and generated images. However, GAN-based training can be complex, prompting some studies to explore non-adversarial approaches for image generation, such as using generative models to create source-like samples \cite{13liu2021source}, or employing prompt learning for style compensation in either the spatial \cite{9hu2022prosfda} or frequency domain \cite{11wang2023fvp}.

Most SFDA approaches combine both data generation and model fine-tuning in a two-stage process: stage one generates target-style images, and stage two uses these images for target model fine-tuning, which may be supervised by methods such as BN statistical information loss \cite{9hu2022prosfda,10yang2022FSM, 12kbshong2022source, 13liu2021source}, pseudo-label loss, or other self-supervised losses \cite{11wang2023fvp, 15chen2021source}. For example, FSM \cite{10yang2022FSM} uses BN statistic loss from both shallow and deep layers to generate images that combine the source domain style and target domain content, which are then used for target model training with a compact-aware consistency module and feature-level contrastive learning. In this study, we combine data generation and model fine-tuning in an end-to-end framework, addressing the domain gap in both the initial and subsequent training phases. We achieve this by leveraging a preadapted model, data-dependent prompt learning, and pseudo-labeling-based style-related layer fine-tuning strategies.

\subsection{Prompt learning}
Prompt learning was originally applied to fine-tuning large language models for downstream tasks, and more recently, visual prompts have been proposed for computer vision tasks \cite{23jia2022visual}. By introducing a visual prompt at the input level, fine-tuning can be achieved by training only a small number of learnable parameters in the prompt while keeping the model's backbone frozen. This process of adapting a large model to a specific downstream task mirrors the process of adapting a source model to a target domain. As a result, prompt learning has been increasingly used for DA in classification \cite{24gao2022visual,25oh2023blackvip} and segmentation tasks \cite{9hu2022prosfda, 11wang2023fvp, 26yang2023exploring}, offering a novel approach for both style translation and model fine-tuning. For example, ProSFDA \cite{9hu2022prosfda} trains a spatial prompt using BN layer statistical loss, which translates target domain images into the source domain style during the first stage of the method. Additionally, some studies have explored training visual prompts in the frequency domain. For instance, FVP \cite{11wang2023fvp} trains a frequency domain prompt through pseudo-label learning while freezing other parameters in the target model. This approach not only achieves style translation of the target domain images but also improves the performance of the target model. While prompt learning has yielded impressive results in DA, existing studies generally treat prompts as domain-dependent parameters, overlooking intrasample differences within the target domain. To address this limitation, we introduce DDFP in this study. 

\subsection{BN statistic calibration}
BN statistic calibration is commonly used in test-time adaptation (TTA) \cite{19li2016revisiting, 27niloy2024source, 28wimpff2024calibration, 29wu2024test,30zhang2023domainadaptor} to recalibrate the batch statistics in the source model using target domain data, thereby making the model more suitable for the target domain. Given that the BN layers play a critical role in model performance under domain gaps \cite{31schneider2020improving} previous studies have shown that adapting only the BN statistics from the target domain to the source model is an effective way to bridge the domain gap. For example, AdaBN \cite{19li2016revisiting} computes a target domain-specific BN statistic at test-time using the entire target dataset, improving the source model's performance on target domain data. Zhang et al. \cite{30zhang2023domainadaptor} argue that directly replacing the source BN statistics with the target domain statistics can lead to performance degradation. To address this, they propose AdaMixBN, which dynamically fuses source and target statistics for TTA. These approaches all use BN calibration to adapt the source model to the target domain data without requiring further training.

While these SFDA methods can reduce the domain gap during training to some extent, a gap still remains between the initialized target model and the target domain data during the initial training phase. As a result, initializing the target model with the source model or generating pseudo-labels for target domain data can introduce errors, leading to performance degradation. To tackle this issue, we propose integrating BN calibration into SFDA as a preadaptation step. This process involves preadapting the source model to an intermediate model (the preadapted model), which provides a more suitable initialization for the target model and improves pseudo-label quality for target domain images. Unlike previous studies, our approach uses BN calibration to reduce the domain gap specifically in the initial training phase. By combining BN calibration with prompt learning, we achieve a more comprehensive reduction of the domain gap across various phases and perspectives. Furthermore, we focus on leveraging BN calibration to enhance pseudo-label quality, which in turn improves the performance of self-supervised learning in SFDA.

\begin{figure*}[!t]
  \centering
  \includegraphics[scale=0.575]{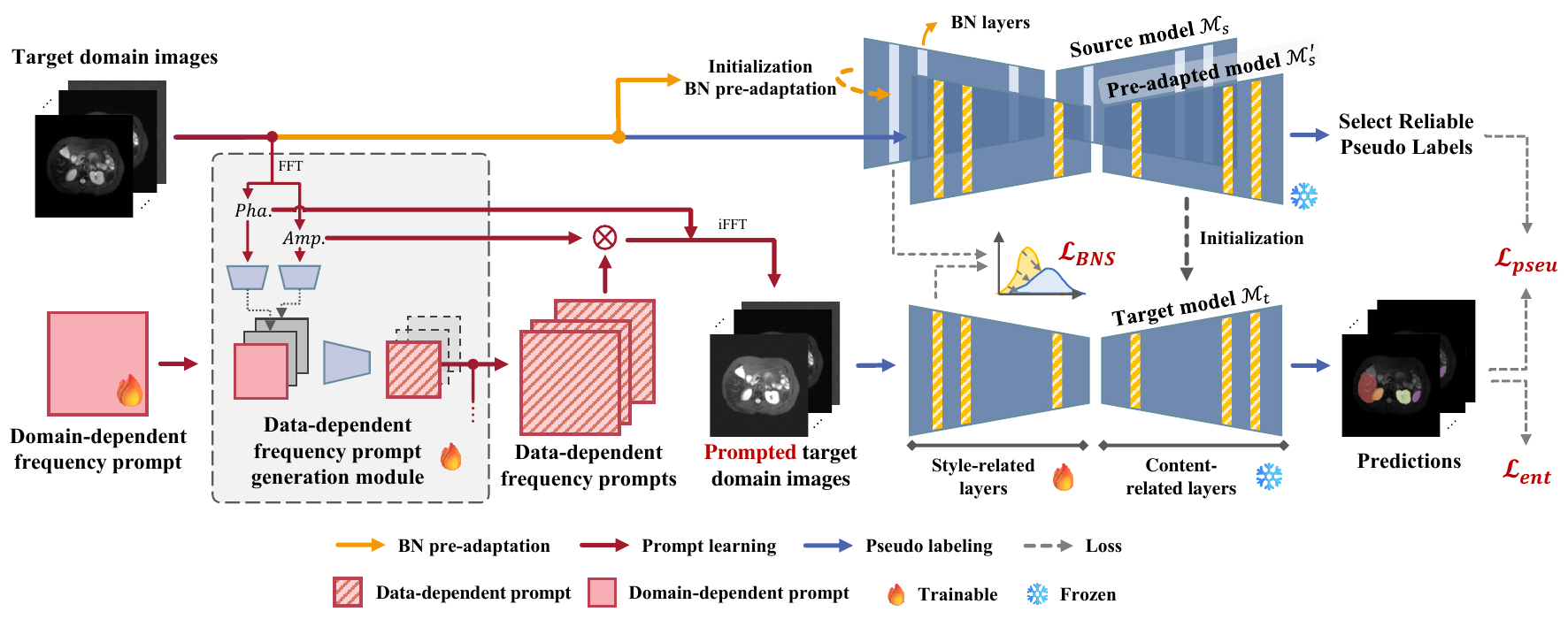}
  \caption{
    Overview of the proposed DDFP architecture. We introduce a BN preadaptation strategy (\textcolor[RGB]{243, 153, 7}{yellow}) to initialize the target model and generate high-quality pseudo-labels for target domain data (\textcolor[RGB]{74, 100, 174}{blue}). The data-dependent frequency prompt learning strategy (\textcolor[RGB]{162, 26, 45}{red}) facilitates image style translation. Both the data-dependent frequency prompt parameters and the style-related layers of the target model are jointly trained to achieve DA.}
  \label{fig2}
\end{figure*}

\section{Methodology}
\subsection{Problem definition}
Let the source domain dataset be $\mathcal{D}_s = \{{x}_j^s,{y}_j^s \}_{j=1}^{N_s}$, which contains $N_s$ samples, where ${x}_j^s\in\mathbb{R} ^{H \times W \times C}$ represents the $j^{th}$ image and ${y}_j^s\in\mathbb{R}^{H \times W \times N_c}$ denotes its segmentation label. $C$ is the number of image channels, and $N_c$ is the number of classes. Similarly, let the target domain dataset be $\mathcal{D}_t=\{{x}_i^t \}_{i=1}^{N_t}$, which contains $N_t$ target domain images, where each ${x}_i^t$ is and image from the target domain. The source model $\mathcal{M}_s$ is initially trained on the source domain dataset $\mathcal{D}_s$. However, due to the domain gap, directly applying $\mathcal{M}_s$ to the target domain data results in performance degradation. Therefore, within the framework of SFDA, our goal is to adapt the knowledge learned by $\mathcal{M}_s$ to the target model $\mathcal{M}_t$ using only the source model $\mathcal{M}_s$ and the unlabeled target domain data $\mathcal{D}_t$ for training. 

\subsection{Overall framework}
To address the DA problem without requiring access to source domain data, we propose a novel SFDA framework named DDFP, as illustrated in Fig. \ref{fig2}. Our goal is to reduce the domain gap throughout both the initial and subsequent training phases by leveraging the preadapted model, data-dependent prompt learning, and pseudo-labeling-based style-related layer fine-tuning.

We begin by applying a BN layer preadaptation strategy to partially calibrate the BN statistics of the source model $\mathcal{M}_s$  using target domain images. This results in the preadapted model $\mathcal{M}_s^{'}$ (depicted by the yellow line in Fig. \ref{fig2}), which is used to initialize the target model $\mathcal{M}_t$ and to generate pseudo-labels for the target domain data.

After initializing the target model with the preadapted model, we proceed with training the target model, focusing exclusively on its style-related layers and prompt-related parameters. At the input level, we use a data-dependent frequency promptDFFP to translate the original target domain images into source-like images, as shown by the red line in Fig. \ref{fig2}. Specifically, we introduce a data-dependent frequency promptDFFP generation module $\mathbf{G_{DDFP}}$, which takes two inputs: the trainable domain-dependent frequency prompt ${FP}^{domain}\in\mathbb{R}^{H \times W}$ and the frequency spectrum of a target domain image. The model outputs the data-dependent frequency prompt $\widetilde{{FP}}_{t,i}^{data}\in\mathbb{R}^{H \times W }$ is then applied to the image amplitude spectrum. An inverse fast Fourier transform (FFT) is used to reconstruct the prompted target domain image.

The prompted images are fed into the target model to generate predictions, which are then supervised using pseudo-labels. To generate the pseudo-labels, the original target domain images are passed through the preadapted model $\mathcal{M}_s^{'}$, producing initial pseudo-labels. Reliable regions in these pseudo-labels are selected to supervise the training of the target model (represented by the upper blue branch in Fig. \ref{fig2}). We compute the Dice loss between the reliable pseudo-label regions and the output of$\mathcal{M}_t$ (represented by the lower blue branch in Fig. \ref{fig2}) for the corresponding prompted target domain images. Additionally, BN statistic loss is calculated between the BN statistics of the source model and the target model to align the style of the prompted images with the source domain images. Finally, entropy loss derived from the target model's predictions is used to jointly supervise the training of both the prompt-related parameters and style-related layers in the target model.

Details of the BN preadaptation strategy are presented in Section 3.3. The strategies for pseudo-label learning and DFFP are introduced in Sections 3.4 and 3.5, respectively. Finally, Section 3.6 outlines the full model training process and the associated loss functions.

\begin{figure*}[!t]
  \centering
  \includegraphics[scale=0.56]{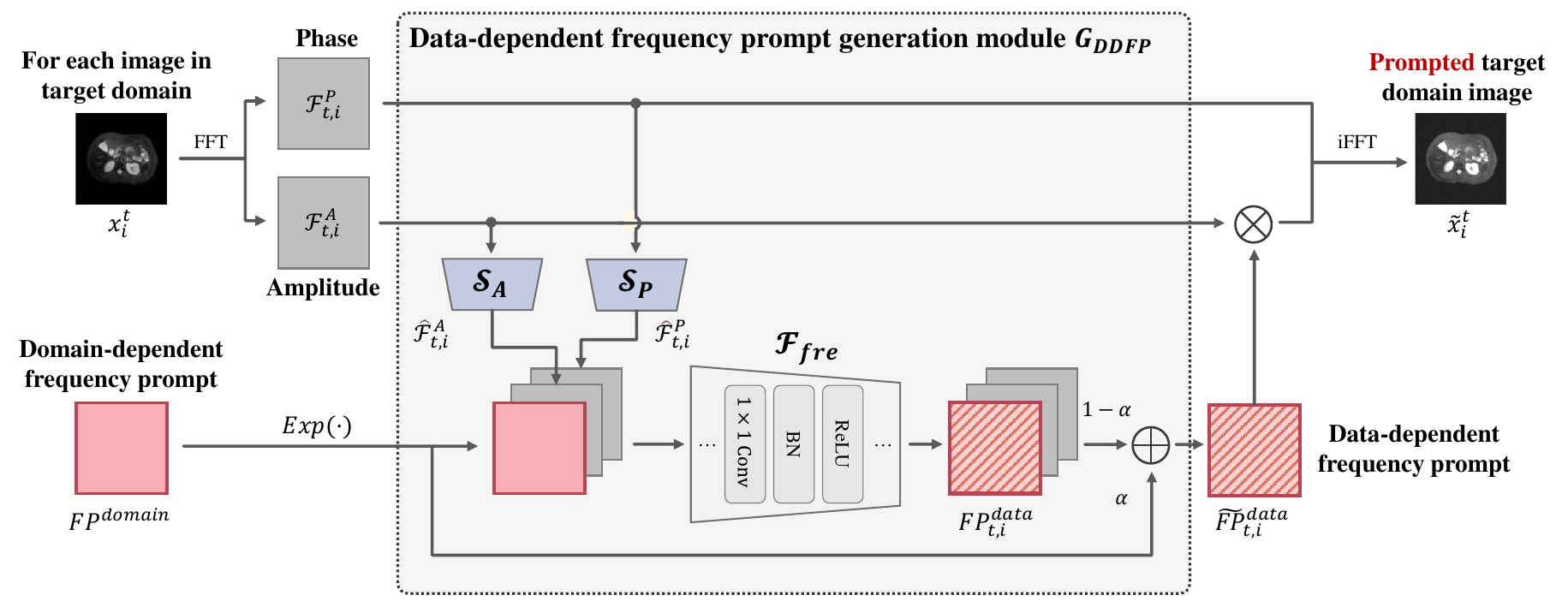}
  \caption{{{Data-dependent frequency prompt generation process for each image in the target domain batch.}}}
  \label{fig3}
\end{figure*}

\subsection{BN pre-adaptation}

Instead of directly using the source model to initialize the target model, we first perform BN preadaptation by recalculating the BN statistics in the source model using target domain images. This process yields the preadapted model $\mathcal{M}_s^{'}$. Inspired by BN calibration methods such as AdaBN \cite{19li2016revisiting}, the preadapted model is obtained by updating the running mean and variance of the source model through a momentum-based approach over $E_W$ epochs of forward propagation, without the need for model parameters or loss function backpropagation. Specifically, the BN statistics of the $l^{th}$ BN layer in the $e^{th}$ iteration are updated as follows:

\begin{equation}
  \begin{aligned}
      \hat{\mu}_l^e & = (1-\rho) \cdot  \hat{\mu}_l^{e-1} + \rho \cdot {\mu}_l^{e, target}, \\
      (\hat{\sigma}_l^e)^2&  =  (1-\rho) \cdot  (\hat{\sigma}_l^{e-1})^2 + \rho \cdot (\sigma_l^{e, target})^2
  \end{aligned}
  \label{eq1}
\end{equation}

where $ \hat{\mu}_l^e$, $(\hat{\sigma}_l^e)^2$ represent the updated BN statistics in the preadapted model $\mathcal{M}_s^{'}$. We use the BN statistics of the source model to initialize $\hat{\mu}_l^0, (\hat{\sigma}_l^0)^2$. ${\mu}_l^{e, target}, (\sigma_l^{e, target})^2$  represent the mean and variance of the current batch of target domain images at the $l^{th}$ BN layer. $\rho$ is the coefficient that controls the mixing of the source and target domains statistic. Once the BN statistics are adapted, the preadapted model $\mathcal{M}_s^{'}$ is used to initialize the target model and generate pseudo-labels for the target domain images.

\subsection{Data-dependent frequency prompt}
To mitigate the domain gap at the image level, we propose using prompted target domain images, rather than the original target domain images, to fine-tune the initialized target model. Previous prompt-based image style translation methods typically rely on domain-dependent prompts, which capture only the interdomain transformation relationships. In contrast, we introduce data-dependent frequency prompts, which also account for intraclass variations.

Figure \ref{fig3} illustrates the process of generating a DFFP for each image in the target domain batch. Given a target domain input image  $x_i^t$, we sequentially process each channel of the image and first perform an FFT to obtain its amplitude and phase spectra $\mathcal{F}_{t,i}^A \in \mathbb{R}^{H \times W}, \mathcal{F}_{t,i}^F \in \mathbb{R}^{H \times W}$.These spectra are then passed into the DFFP generation module $\mathbf{G_{DDFP}}$, along with a trainable domain-dependent frequency prompt ${FP}^{domain}\in\mathbb{R}^{H \times W}$. In $\mathbf{G_{DDFP}}$, $\mathcal{F}_{t,i}^A, \mathcal{F}_{t,i}^F$ are processed by two separate simple neural networks $\mathbf{\mathcal{S}_A}, \mathbf{\mathcal{S}_P}$, respectively. The output features are then concatenated along the channel dimension with ${FP}^{domain}$, and the resulting feature map is processed by another neural network $\mathbf{ \mathbf{\mathcal{F}_{fre}}}$. $\mathbf{\mathcal{S}_A}, \mathbf{\mathcal{S}_P}$ and $\mathbf{ \mathbf{\mathcal{F}_{fre}}}$ within the $\mathbf{G_{DDFP}}$ are simple networks composed of 1 $\times$ 1 convolution layers, ReLU activations, and other basic activation layers. We assign the output channel corresponding to ${FP}^{domain}$ of $\mathbf{ \mathbf{\mathcal{F}_{fre}}}$ as ${FP}_{t,i}^{data}$. The final data-dependent frequency prompt $\widetilde{{FP}}_{t,i}^{data}$ is obtained through a skip connection between ${FP}^{domain}$ and ${FP}_{t,i}^{data}$. $\widetilde{{FP}}_{t,i}^{data}$ is then multiplied with the amplitude spectrum of $x_i^t$ and passed through the inverse FFT, along with the original phase spectrum, to reconstruct the prompted target domain image $\widetilde{x}_i^t$ in the spatial domain. 

Specifically, $\mathbf{\mathcal{S}_A} and \mathbf{\mathcal{S}_P}$ consist of two sets of 1 $\times$ 1 convolution layer, BN layer and ReLU activations to preprocess $\mathcal{F}_{t,i}^A, \mathcal{F}_{t,i}^F$.
\begin{equation}
  \hat{\mathcal{F}}_{t,i}^A = \mathbf{\mathcal{S}_A}(\mathcal{F}_{t,i}^A), \hat{\mathcal{F}}_{t,i}^F =\mathbf{\mathcal{S}_P}(\mathcal{F}_{t,i}^F).
  \label{eq2}
\end{equation}

Next, $\hat{\mathcal{F}}_{t,i}^A \in \mathbb{R}^{H \times W}, \hat{\mathcal{F}}_{t,i}^F \in \mathbb{R}^{H \times W}$ and ${FP}^{domain}$ are concatenated along the channel dimension and passed through $\mathbf{\mathcal{F}_{fre}}$. {{$\hat{\mathcal{F}}_{t,i}^A,  \hat{\mathcal{F}}_{t,i}^F$ help ${FP}^{domain}$ learn the specific variance of each image, thereby adapting ${FP}^{domain}$ into the corresponding ${FP}^{data}_{t,i}$.}} $\mathbf{\mathcal{F}_{fre}}$ consists of three sets of 1 $\times$ 1 convolution layers, BN layers, and ReLU activations, facilitating the interaction between the prompt and the frequency spectra. Ultimately, the channel corresponding to ${FP}^{domain}$ in the output of $\mathbf{\mathcal{F}_{fre}}$ is extracted as ${FP}_{t,i}^{data}$.
\begin{equation}
    {FP}_{t,i}^{data} = \mathbf{\mathcal{F}_{fre}} \biggl( cat (\hat{\mathcal{F}}_{t,i}^A,  \hat{\mathcal{F}}_{t,i}^F, Exp({FP}^{domain})) \biggr)[2,...]
    \label{eq3}
\end{equation}
where ${cat}$ refers to the concatenation operation along the channel dimension, resulting in a matrix of $\mathbb{R}^{3 \times H \times W}$. The $Exp(\cdot)$ operation ensures that the domain-dependent frequency prompt maintains non-negative values, similar to the other two spectrum components. The notation $[2,...]$ indicates that the last channel in the output of $\mathbf{\mathcal{F}_{fre}}$ is taken as ${FP}_{t,i}^{data}$ (count from zero), corresponding to the dimension of ${FP}^{domain}$. We use a skip connection betweenn ${FP}^{domain}$ and ${FP}_{t,i}^{data}$ to obtain the final data-dependent frequency prompt  $\widetilde{FP}_{t,i}^{data}$.
\begin{equation}
  \widetilde{{FP}}_{t,i}^{data} = \alpha \times Exp({FP}^{domain}) + (1-\alpha) \times {FP}_{t,i}^{data}
  \label{eq4}
\end{equation}
where $\alpha$ is the fusion weight. Given that the ideal prompt aims to achieve style translation without altering the content, we apply $ \widetilde{{FP}}_{t,i}^{data}$ on the amplitude spectrum $\mathcal{F}_{t,i}^A$ and then use inverse $F^{-1}(\cdot)$ to obtain the final prompted target image $\widetilde{x}_i^t$.
\begin{equation}
    \widetilde{x}_i^t = F^{-1} (\mathcal{F}_{t,i}^A \odot \widetilde{{FP}}_{t,i}^{data}, \mathcal{F}_{t,i}^P )
    \label{eq5}
\end{equation}
where $\odot$ denotes the element-wise multiplication operator.

\subsection{Pseudo labeling}
Instead of directly using the original source model to generate pseudo-labels for target domain images, we utilize the predictions from the preadapted model $\mathcal{M}_s^{'}$ as initial pseudo-labels for the target domain. These preliminary pseudo-labels are then refined through filtering based on category and global thresholds to retain only the most reliable labels. Finally, the pseudo-labeling loss is computed between the refined pseudo-labels and the output of the target model, with pixel-wise confidence weights to adjust the loss according to the reliability of each pixel.

For each target domain image $x_i^t$, the prediction from $\mathcal{M}_s^{'}$ is denoted as $p^{\mathcal{M}_s^{'}}(x_i^t) \in \mathbb{R}^{H \times W \times N_c}$. The preliminary one-hot pseudo label $\hat{y}^{\mathcal{M}_s^{'}}(x_i^t) \in \mathbb{R}^{H \times W \times N_c}$ is assigned based on the class with the highest probability for each pixel. To assess the reliability of these pseudo-labels, we calculate the pixel-wise entropy ${ent}_{h,w}^{\mathcal{M}_s^{'}}(x_i^t)$ for each pixel. Pixels with entropy values below two predefined thresholds are considered reliable and are used to form the refined pseudo-labels $\widetilde{y}^{\mathcal{M}_s^{'}}_{h,w}(x_i^t)$. Specifically, the entropy is computed in Equ. \eqref{eq6}.
\begin{equation}
  {ent}_{h,w}^{\mathcal{M}_s^{'}}(x_i^t) = - \sum_{c=1}^{N_c} (p_{c,h,w}^{\mathcal{M}_s^{'}}(x_i^t) log(p_{c,h,w}^{\mathcal{M}_s^{'}}(x_i^t)))
  \label{eq6}
\end{equation}
where $p_{c,h,w}^{\mathcal{M}_s^{'}}(x_i^t)$ denotes the predicted probability of pixel $(h,w)$ for class $c$.

We use a set of category-specific entropy thresholds $\delta_{cls} = \{ t_{cls0}, \ldots ,t_{cls{N_c}} \}$ to filter out unreliable pixels in each class. Here, $t_{clsc} \in [0,1]$ represents the proportion of pixels to be retained for class $c$. {{This threshold ensures that only reliable pixels are used for loss calculation in each class, preventing situations where background pixels (which are more abundant and are easier to classify with smaller entropy values) dominate the reliable pseudo-labels.}} The value $\tau$ represents the entropy value corresponding to the $\tau(t_{clsc})$-quantile pixels. A category-level reliable pixel is then defined as follows:
\begin{equation}
    Cls \_ \widetilde{y}_{c,h,w}^{\mathcal{M}_s^{'}}(x_i^t) =  \mathbb{I}  \left[ \hat{y}_{c,h,w}^{\mathcal{M}_s^{'}}(x_i^t) = 1 ~ {and} ~ {ent}_{h,w}^{\mathcal{M}_s^{'}}(x_i^t) < \tau(t_{cls\_c}) \right]
  \label{eq7}
\end{equation}
where $\mathbb{I}$ is the indicator function. {{In addition,when the domain gap is large, pixels that are filtered out by $\delta_{cls}$ may still have high entropy but are mistakenly identified as reliable pseudo-labels.}} To address this, we introduce a global entropy threshold $\delta_{glo}$, which helps further filter out such unreliable pixels based on their overall entropy values. This ensures that the remaining reliable labels not only have lower entropy within their respective classes but also possess globally lower entropy, making them more trustworthy.

\begin{equation}
    Glo \_ \widetilde{y}_{h,w}^{\mathcal{M}_s^{'}}(x_i^t) = 
     \mathbb{I} \left[ {ent}_{h,w}^{\mathcal{M}_s^{'}}(x_i^t) < \delta_{glo} \right]
  \label{eq8}
\end{equation}

The final selection of reliable labels $\widetilde{y}_{h,w}^{\mathcal{M}_s^{'}}(x_i^t)$ is carried out as follows:
\begin{equation}
      \widetilde{y}_{c,h,w}^{\mathcal{M}_s^{'}}(x_i^t) = 
       \mathbb{I} \left[ Cls \_ \widetilde{y}_{h,w}^{\mathcal{M}_s^{'}}(x_i^t) \right]
     \mathbb{I}  \left[ Glo \_ \widetilde{y}_{h,w}^{\mathcal{M}_s^{'}}(x_i^t) \right] 
  \label{eq9}
\end{equation}

\subsection{Target model fine-tuning and loss function}
Instead of fine-tuning the entire target model, we only update the parameters of the shallow, style-related layers and freeze the deep, content-related layers. Given that there is no clear-cut distinction between style and content layers, we designate the first four convolutional layers of our U-Net backbone \cite{33ronneberger2015u} as the style-related layers, which are trainable. The remaining layers are treated as content-related and are frozen during the fine-tuning process (as shown in the bottom-right part of Fig. \ref{fig2}).

Both the DFFP parameters and the style-related layers in the target model are trained simultaneously. The goal of using the data-dependent frequency promptDFFP is twofold: (i) the data distribution of the prompted images should match that of the source domain images, and (ii) the model output should closely resemble the one-hot labels, exhibiting minimal entropy. To achieve this, we introduce two loss functions. The first is the BN statistic loss ($\mathcal{L}_{BNS}$), which calculates the discrepancy between the statistical metrics (mean and variance) of the source model's BN layers and those of the target model, aligning the style of the prompted target domain images with the source domain images. The loss is defined as follows:

\begin{equation}
  \mathcal{L}_{BNS} = \sum_{l=0}^L ({\parallel \mu_l^{\mathcal{M}_s} -\mu_l^{\mathcal{M}_t} \parallel}_2 + {\parallel (\sigma_l^{\mathcal{M}_s})^2 -(\sigma_l^{\mathcal{M}_t})^2 \parallel}_2)
  \label{eq10}
\end{equation}
where ${\parallel \cdot \parallel}_2$ denotes the $\mathcal{L}$2-norm. In addition, we apply an entropy minimization loss to supervise the model at the output level:
\begin{equation}
  \mathcal{L}_{ent} = - \frac{1}{H \times W} \sum_{h}^{H} \sum_{w}^{W} p_{h,w}^{\mathcal{M}_t}(x_i^t) log( p_{h,w}^{\mathcal{M}_t}(x_i^t))
  \label{eq11}
\end{equation}

The prompted target domain images are passed through $\mathcal{M}_t$, and the model's predictions are denoted as $p^{\mathcal{M}_t}(x_i^t)$. The selected pseudo-labels $ \widetilde{y}^{\mathcal{M}_s^{'}}(x_i^t)$ which are derived from the preadapted model, are then used to supervise the model training by calculating the cross-entropy loss. The loss is reweighted by the pixel-wise prediction confidence ${conf}_{h,w}^{\mathcal{M}_s^{'}}(x_i^t)$, which is determined by the maximum prediction probability for each pixel.
\begin{equation}
  \begin{aligned}
      \mathcal{L}_{pseu} &= - \frac{\vartheta }{\theta} \sum_{h}^{H} \sum_{w}^{W} 
      [\widetilde{y}_{h,w}^{\mathcal{M}_s^{'}}(x_i^t) log(p_{h,w}^{\mathcal{M}_t}(x_i^t)) \\
      &+ (1- \widetilde {y}_{h,w}^{\mathcal{M}_s^{'}}(x_i^t)) log(1-p_{h,w}^{\mathcal{M}_t}(x_i^t)) ]   {conf}_{h,w}^{\mathcal{M}_t}(x_i^t)  \\
       \theta &= (H \times W) / (|\widetilde{y}_{h,w}^{\mathcal{M}_s^{'}}(x_i^t)|)
  \end{aligned}
  \label{eq12}
\end{equation}
where $\theta$ is the parameter for regulation. $\vartheta$ is a hyperparameter. $|\widetilde{y}_{h,w}^{\mathcal{M}_s^{'}}(x_i^t)|$ is the count of selected reliable pseudo-label pixels. Unreliable pixels that do not meet the criteria defined in Eq. \eqref{eq3} are excluded from the loss calculation.

Finally, we use the $\mathcal{L}_{total}$ to fine-tune the style-related layers and train the prompt-related parameters in the target model, with $w_{ent}, w_{BNS}, w_{pseu}$ representing the weights associated to each loss component.
\begin{equation}
    \mathcal{L}_{total} = w_{ent} \times \mathcal{L}_{ent} + w_{BNS} \times  \mathcal{L}_{BNS} +  w_{pseu} \times \mathcal{L}_{pseu}
    \label{eq14}
\end{equation}

\begin{table*}[htbp]
  \centering
  \setlength{\tabcolsep}{1.1mm}
  \caption{Quantitative segmentation results on the multiorgan abdominal dataset. The best results are highlighted in bold, and the second-best results are underlined.}
    \begin{tabular}{ccccccccccccc}
    \toprule
    \multicolumn{13}{c}{Abdominal } \\
    \midrule
    \multirow{2}[4]{*}{{{Backbone}}} & \multirow{2}[4]{*}{\makecell{Method \\ (CT→MRI)}} & \multicolumn{5}{c}{Dice ↑}            &       & \multicolumn{5}{c}{ASD (mm) ↓} \\
\cmidrule{3-7}\cmidrule{9-13}          &       & Liver & R.kidney & L.kidney & Spleen & Average &       & Liver & R.kidney & L.kidney & Spleen & Average \\
    \midrule
    \multirow{5}[4]{*}{{{U-Net}}} & Supervised & 0.9555 & 0.9532 & 0.9470 & 0.9346 & 0.9476 &       & 0.6876 & 0.9013 & 0.6324 & 1.0924 & 0.8284 \\
          & W/o adaptation & 0.5640 & 0.8655 & 0.8464 & 0.4118 & 0.6719 &       & 2.6498 & 0.9067 & 0.6387 & 4.5344 & 2.1824 \\
          \cmidrule{2-13}  
         & ProContra\cite{14yu2023source} & \underline{0.7933} & \underline{0.9132} &\underline{0.8824} & \underline{0.7641} &\underline{0.8382} &       & \textbf{0.3296} & 3.1929 & 3.8131 & \textbf{1.7593} & \underline{2.2737} \\
          & TT-SFUDA\cite{48ttsfuda} & 0.7284 & 0.7806 & 0.8561 & 0.4690 & 0.7085 &       & 1.9631 & \underline{2.6570} & \underline{0.5707} & 4.8854 & 2.5191 \\
          & {{DDFP (ours)}} & {{\textbf{0.9053 }}} & {{\textbf{0.9206 }}} & {{\textbf{0.9263 }}} & {{\textbf{0.8426 }}} & {{\textbf{0.8987 }}} &       & {{\underline{0.8336}}}  & {{\textbf{0.3426 }}} & {{\textbf{0.4554 }}} & {{\underline{4.5091}}}  & {{\textbf{1.5352 }}} \\
    \midrule
    \multirow{7}[6]{*}{{{DeepLab}}} & {{Supervised}} & {{0.9364}} & {{0.9520}} & {{0.9371}} & {{0.9294}} & {{0.9387}} &       & {{0.4884}} &{{ 0.1350 }}& {{0.2885}} & {{0.3240}} & {{0.3090}} \\
     & {{W/o adaptation}} & {{0.7614}} &{{ 0.8695}} & {{0.7740}} &{{ 0.6214}} & {{0.756}}6 &       & {{2.1741}} & {{1.2733}} & {{1.3547}} & {{2.1500}} & {{1.7380}} \\
     \cmidrule{2-13}  
      & DPL\cite{15chen2021source}   & \textbf{0.8775} & 0.7860 & 0.5779 & \underline{0.7733} & \underline{0.7537} &       & \textbf{1.4094} & 3.2024 & 2.2696 & \underline{1.6145} & \underline{2.1240} \\
          & CBMT\cite{47tang2023source}  & \underline{0.8431} & 0.5311 & 0.7619 & 0.7652 & 0.7253 &       & \underline{1.7181} & 3.0600 & 6.6043 & 2.3374 & 3.4300 \\
          & FSM\cite{10yang2022FSM}*  & 0.6320 & 0.8540 & 0.7960 & 0.5080 & 0.6975 &       & 4.7700 & \underline{2.5460} & 1.7210 & 6.7550 & 3.9480 \\
          & FVP\cite{11wang2023fvp}*  & 0.6480 & \underline{0.8760} & \underline{0.8030} & 0.6050 & 0.7330 &       & 4.4830 & \textbf{2.1010} & \underline{1.5420} & 6.1530 & 3.5698 \\
          & {{DDFP (ours)}} & {{0.7806}} & {{\textbf{0.8927}}} & {{\textbf{0.8747}}} & {{\textbf{0.8522}}} & {{\textbf{0.8501}}} &       & {{1.8730}} &{{ 2.5598}} & {{\textbf{0.7789}}} & {{\textbf{1.3124}}} &{{ \textbf{1.6311}}} \\
    \midrule
    \midrule
    \multirow{2}[4]{*}{Backbone} & \multirow{2}[4]{*}{\makecell{Method \\ (MRI→CT)}} & \multicolumn{5}{c}{Dice↑}             &       & \multicolumn{5}{c}{ASD (mm) ↓} \\
\cmidrule{3-7}\cmidrule{9-13}          &       & Liver & R.kidney & L.kidney & Spleen & Average &       & Liver & R.kidney & L.kidney & Spleen & Average \\
    \midrule
    \multirow{5}[4]{*}{{{U-Net}}} & Supervised & 0.9528 & 0.9112 & 0.9064 & 0.9369 & 0.9268 &       & 0.6876 & 0.9013 & 0.6324 & 1.0924 & 0.8284 \\
          & W/o adaptation & 0.6198 & 0.3873 & 0.3541 & 0.5453 & 0.4766 &       & 4.8115 & 16.6979 & 10.3214 & 6.7980 & 9.6572 \\
\cmidrule{2-13}          & ProContra\cite{14yu2023source} & \textbf{0.8741} & \underline{0.6864} & \underline{0.7274} & \underline{0.7014} & \underline{0.7473} &       & \textbf{1.9015} & \textbf{8.9773} & 7.3395 & 6.3545 & 8.2318 \\
          & TT-SFUDA\cite{48ttsfuda} & 0.8473 & 0.4954 & 0.6837 & 0.7009 & 0.6818 &       & 3.3714 & 16.9375 & \underline{6.7724} & \underline{4.8754} & \underline{7.9891} \\
          & {{DDFP (ours)}} & {{\underline{0.8623}}}  &{{ \textbf{0.7386 }}} & {{\textbf{0.7746 }}} & {{\textbf{0.7980}} } & {{\textbf{0.7934 }}} &       & {{\underline{2.3012}}}  &{{\underline{10.8627}}}  & {{\textbf{4.6034 }}} & {{\textbf{3.3268 }}} & {{\textbf{5.2735 }}} \\
    \midrule
    \multirow{7}[2]{*}{{{DeepLab}}} & {{Supervised}} & {{0.9536}} & {{0.9122}} & {{0.8949}} & {{0.9246}} & {{0.9213}} &       & {{0.6281}} & {{0.5225}} & {{0.7111}} & {{0.6156}} & {{0.6193}} \\
          & {{W/o adaptation}} & {{0.3629}} & {{0.5075}} &{{ 0.4361}} & {{0.5453}} & {{0.4630}} &       & {{9.2460}} & {{7.9282}} & {{5.5497}} & {{9.1842}} & {{7.9770}} \\
          \cmidrule{2-13}  
          & DPL\cite{15chen2021source}   & 0.7674 & 0.5340 & 0.5846 & 0.7060 & 0.6480 &       & 5.5458 & 10.9472 & 12.1929 & 8.1928 & 9.2197 \\
          & CBMT\cite{47tang2023source}  & \textbf{0.9008} & \underline{0.6811} & 0.6666 & \textbf{0.7539} & \underline{0.7506} &       & \textbf{2.3692} & 9.5435 & 12.6072 & 5.1747 & 7.4237 \\
          & FSM\cite{10yang2022FSM}*  & 0.8700 & 0.6190 & 0.6940 & 0.6880 & 0.7178 &       & 4.5840 & 4.6960 & 3.9020 & \underline{4.1130} & 4.3238 \\
          & FVP\cite{11wang2023fvp}*  & \underline{0.8780} & 0.6470 & \underline{0.7320} & 0.6830 & 0.7350 &       & 3.6310 & \textbf{2.5830} & 3.1020 & \textbf{2.3360} & \textbf{2.9130} \\
          & {{DDFP (ours)}} & {{0.8163}} & {{\textbf{0.8055}}} & {{\textbf{0.7535}}} & {{\underline{0.7215}}} & {{\textbf{0.7742}}} &       & {{\underline{3.2415}}} & {{\underline{4.4681}}} & {{\textbf{2.5175}}} & {{4.7905}} & {{\underline{3.7544}}} \\
    \bottomrule
    \end{tabular}%
  \label{tab:tab1}%
\end{table*}%

\section{Experiments and results}
\subsection{Datasets and experimental setup}

\subsubsection{Datasets}
We evaluated our method and compared it with state-of-the-art methods on two datasets.

\textbf{Multi-organ abdominal dataset.} This dataset consists of 20 MRI volumes from the CHAOS challenge \cite{34kavur2021chaos} and 30 CT volumes from the MICCAI 2015 Multi-Atlas Labeling Beyond the Cranial Vault Workshop and Challenge \cite{35landman2015miccai}. The segmentation labels cover four organs: the liver, left kidney (L. kidney), right kidney (R. kidney), and spleen. We use two-dimensional (2D) slices extracted from the three-dimensional (3D) volumes as separate inputs, discarding slices without labels. CT images are adjusted using a window width and level of [400, 40], while the intensity of MRI images is rescaled to the range of [0, 1200]. All image pixel values are normalized to the range [0, 1], and the images are resized to $256 \times 256$. Within both domains, we randomly split the dataset into training and test sets with an 8:2 ratio. Experiments are conducted for both MRI to CT and CT to MRI adaptation.

\textbf{Cardiac dataset.} This dataset includes 20 MRI volumes and 20 CT volumes from the MMWHS 2017 challenge \cite{36zhuang2016multi}, with segmentation labels for the ascending aorta, left atrium blood cavity, left ventricle blood cavity, and myocardium of the left ventricle. The same preprocessing steps with \cite{40_liu2023reducing,37wang2024towards} are applied. All images are normalized to [0, 1] and resized to $256 \times 256$. We randomly split the dataset into training and test sets with an 8:2 ratio, and experiments are performed for both MRI to CT and CT to MRI adaptation.

{{\textbf{Brain tumor BraTS2018 dataset.} }}
{{This dataset consists of data from 75 patients \cite{49bratsmenze2014multimodal}, including four modalities: T1, T1c, T2, and Flair. We randomly split the dataset into training and test sets with an 8:2 ratio and conduct adaptation between T2 and Flair modalities. The original data includes four labels: Background, necrotic tumor core, peritumoral edema, and enhancing tumor. We combine these labels into two categories: background and foreground tumor areas. All images are normalized to the range [0, 1] and retain their original size of $240 \times 240$.}}

\begin{figure*}[!t]
  \centering
  \includegraphics[scale=0.257]{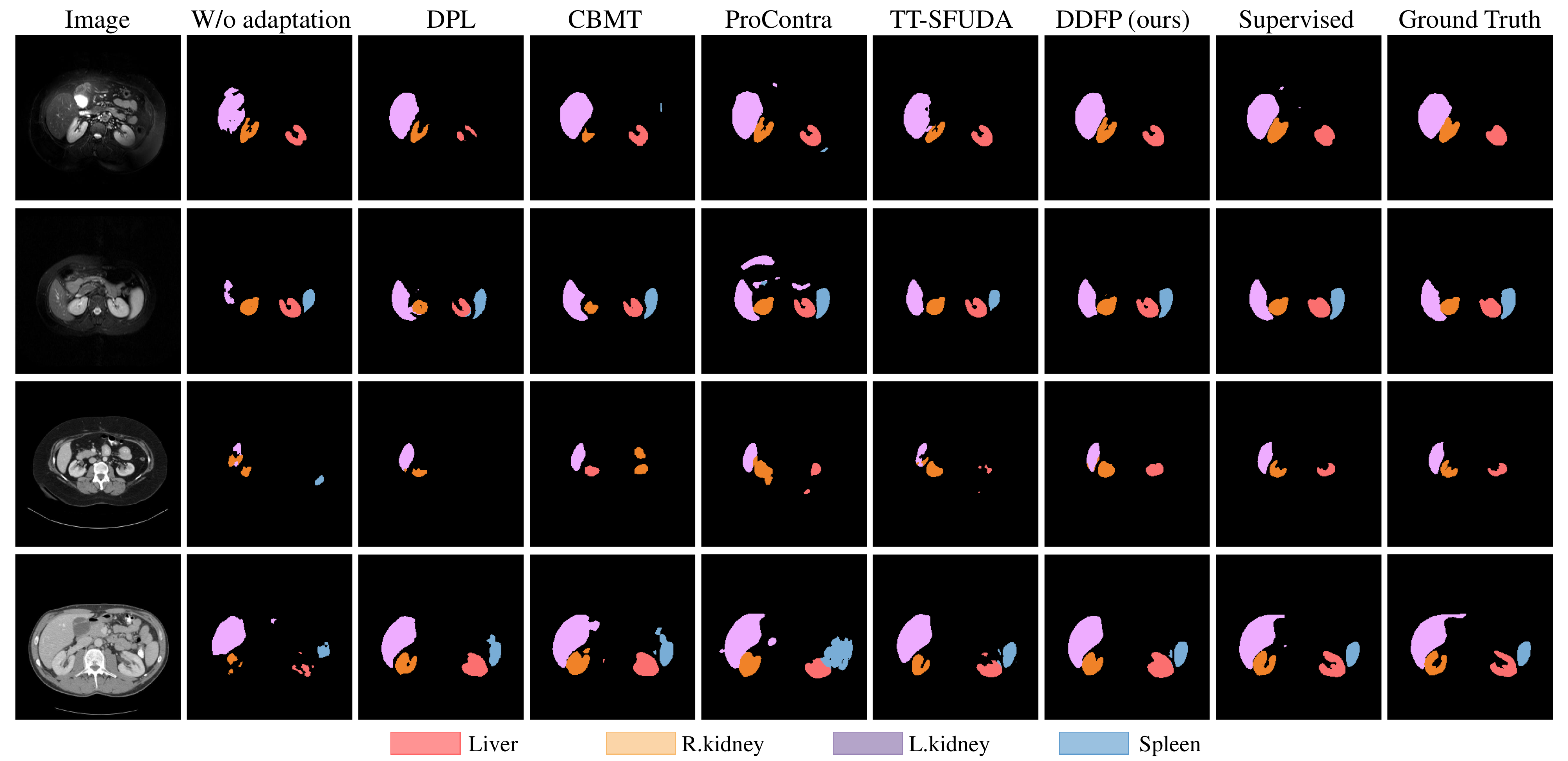}
  \caption{{{Visualization of SFDA segmentation results on the multiorgan abdominal dataset.}} The first two rows show the results for CT to MRI adaptation, while the last two rows display results for MRI to CT adaptation.}
  \label{fig4}
\end{figure*}

\subsubsection{Evaluation metrics}
The model is trained using 2D images, and the final output is reorganized to calculate 3D performance metrics: the Dice coefficient (Dice) and the average symmetric surface distance (ASD). These metrics are consistent with the evaluation methods used in \cite{11wang2023fvp}. The Dice coefficient measures the overlap between the predicted and ground truth labels, with a larger value indicating better model performance. Conversely, ASD evaluates the accuracy of predicted edges, with a smaller value signifying more accurate edge prediction.

\subsubsection{Implementation details}
We use both U-Net \cite{33ronneberger2015u} and DeepLab v3 (with resnet50 backbone) \cite{50deeplabhe2016deep} as the backbone for the model. {{Unless otherwise specified, the same parameter settings are used for different backbones and datasets.}} The source model $\mathcal{M}_s$ is trained on the source domain data using a combination of cross-entropy loss and Dice loss. We optimize the model using the Adam optimizer, with a learning rate of 0.001 for the multiorgan abdominal dataset and 0.0005 for the cardiac dataset. The weight decay is set to 0.0005 for both datasets, and the model is trained for 150 epochs with a batch size of 16.

In the DA stage, we first perform non-training BN preadaptation to obtain the preadapted model $\mathcal{M}_s^{'}$, using $\rho=0.1$ and $E_W=10$. The $\mathcal{M}_s^{'}$ is then used to initialize the target model $\mathcal{M}_t$. {{${FP}^{domain}$ is initialized to zeros.}} We proceed to train the style-related layers in $\mathcal{M}_t$ and the DFFP-related parameters for five epochs. The learning rates for the abdominal and cardiac datasets are set to 0.0005 and 0.001, respectively. The weight decay and batch size are both set to 0.0005 and 16, respectively. The skip connection parameter $\alpha$ in data-dependent frequency prompt learning is set to 0.2. {{$\delta_{cls}$ is set to 40 for all classes,}} and  $\delta_{glo}$ is set to 0.4. $\vartheta$ for the pseudo-labeling loss is set to 0.2.

\begin{table*}[t]
  \centering
  \setlength{\tabcolsep}{1mm}
  \caption{Quantitative segmentation results on the cardiac dataset. The best results are highlighted in bold, and the second-best results are underlined.}
    \begin{tabular}{ccccccccccccc}
    \toprule
    \multicolumn{13}{c}{Cardiac} \\
    \midrule
    \multirow{2}[4]{*}{Backbone} & \multirow{2}[4]{*}{\makecell{Method \\ (CT→MRI)}} & \multicolumn{5}{c}{Dice ↑}            &       & \multicolumn{5}{c}{ASD (mm) ↓} \\
\cmidrule{3-7}\cmidrule{9-13}          &       & AA    & LAV   & LVC   & MYO   & Average &       & AA    & LAV   & LVC   & MYO   & Average \\
    \midrule
    \multirow{5}[4]{*}{ {{U-Net}}} & Supervised & 0.7958 & 0.8499 & 0.9303 & 0.8731 & 0.8623 &       & 1.7941 & 3.1657 & 1.4831 & 2.9046 & 2.3369 \\
          & W/o adaptation & 0.3421 & 0.3457 & 0.6728 & 0.2721 & 0.4082 &       & 12.7709 & 16.6913 & 12.1498 & 13.5387 & 13.7877 \\
\cmidrule{2-13}          & ProContra\cite{14yu2023source} & \textbf{0.6600} & \underline{0.5016} & \underline{0.8252} & 0.5004 & \underline{0.6218} &       & \underline{4.3868} & \underline{13.2456} & \underline{6.0508} & 13.6937 & \underline{9.3442} \\
          & TT-SFUDA\cite{48ttsfuda} & 0.3240 & 0.4009 & 0.7630 & \underline{0.5798} & 0.5169 &       & 14.0603 & 14.8565 & 8.9135 & \underline{12.9439} & 12.6935 \\
          & DDFP (ours) & \underline{0.6499}  & \textbf{0.5712 } & \textbf{0.8384 } & \textbf{0.6907 } & \textbf{0.6876 } &       & \textbf{3.6269 } & \textbf{12.9096 } & \textbf{6.0451 } & \textbf{9.8911 } & \textbf{8.1182 } \\
    \midrule
    \multirow{7}[2]{*}{ {{DeepLab}}} &  {{Supervised }}&  {{0.8203}} &  {{0.8667}} &  {{0.9376}} &  {{0.8613}} &  {{0.8715}} &       &  {{1.2733}} &  {{1.6110 }}&  {{1.0159}} &  {{2.7467}} & {{ 1.6617}} \\
          &  {{W/o adaptation}} &  {{0.5120}} & {{ 0.4341}} &  {{0.6603}} &  {{0.2893}} &  {{0.4739}} &       &  {{4.9264}} &  {{12.0254}} &  {{6.3481}} &  {{8.5610}} &  {{7.9652}} \\
          \cmidrule{2-13}
          & DPL\cite{15chen2021source}   & \underline{0.7261} & \underline{0.6159} & 0.7144 & 0.3755 & \underline{0.6080} &       & \underline{5.7678} & \underline{12.2446} & \underline{10.5101} & \underline{13.7122} & \underline{10.5587} \\
          & CBMT\cite{47tang2023source} & 0.6457 & 0.4450 & \underline{0.7811} & 0.2973 & 0.5423 &       & 10.5237 & 14.3923 & 10.7777 & 19.4049 & 13.7747 \\
          & FSM\cite{10yang2022FSM}*   & 0.5040 & 0.4130 & 0.5170 & 0.4490 & 0.4760 &       & 12.4600 & 27.0920 & 23.7580 & 17.8830 & 20.2983 \\
          & FVP\cite{11wang2023fvp}*   & 0.3850 & 0.4480 & 0.5780 & \underline{0.4910} & 0.4760 &       & 19.0120 & 24.6610 & 18.9230 & 14.5590 & 19.2888 \\
          & DDFP (ours) & \textbf{0.7607} & \textbf{0.8309} & \textbf{0.9028} & \textbf{0.6982} & \textbf{0.7981} &       & \textbf{2.2280} & \textbf{2.2149} & \textbf{2.2343} & \textbf{5.0821} & \textbf{2.9398} \\
    \midrule
    \midrule
    \multirow{2}[4]{*}{Backbone} & \multirow{2}[4]{*}{\makecell{Method \\ (MRI→CT)}} & \multicolumn{5}{c}{Dice ↑}            &       & \multicolumn{5}{c}{ASD (mm) ↓} \\
\cmidrule{3-7}\cmidrule{9-13}          &       & AA    & LAV   & LVC   & MYO   & Average &       & AA    & LAV   & LVC   & MYO   & Average \\
    \midrule
    \multirow{5}[4]{*}{ {{U-Net} }}& Supervised & 0.9010 & 0.9129 & 0.9230 & 0.8648 & 0.9004 &       & 2.4922 & 5.7742 & 3.1959 & 4.6332 & 4.0239 \\
          & W/o adaptation & 0.4061 & 0.8243 & 0.7104 & 0.7413 & 0.6705 &       & 6.6844 & 9.6263 & 7.1081 & 10.4007 & 8.4549 \\
\cmidrule{2-13}          & ProContra\cite{14yu2023source} & 0.5969 & 0.8466 & \underline{0.7742} & 0.7485 & 0.7416 &       & 11.3739 & \underline{7.2578} & 14.0045 & 16.8695 & 12.3765 \\
          & TT-SFUDA\cite{48ttsfuda} & \underline{0.6327} & \underline{0.8758} & 0.6565 & \underline{0.8329} & \underline{0.7495} &       & \underline{5.6244} & 13.5741 & \underline{8.4491} & \underline{12.2613} & \underline{9.9772} \\
          &  DDFP (ours) &  \textbf{0.7088} &  \textbf{0.8923 } &  \textbf{0.8751 } &  \textbf{0.9147 } &  \textbf{0.8477 } &       &  \textbf{4.9193 } &  \textbf{3.6187 } &  \textbf{3.9505 } &  \textbf{5.1714 } &  \textbf{4.4150 } \\
    \midrule
    \multirow{7}[2]{*}{ {{DeepLab}}} &  {{Supervised}} &  {{0.8997}} &  {{0.9225}} &  {{0.9320}} &  {{0.8784}} &  {{0.9081}} &       &  {{1.8698}} &  {{2.7097}} &  {{1.7290}} &  {{2.4737}} &  {{2.1955}} \\
          &  {{W/o adaptation}} &  {{0.7345}} &  {{0.9014}} &  {{0.8768}} &  {{0.8519}} &  {{0.8411}} &       &  {{4.5482}} &  {{5.3496}} &  {{4.3756}} &  {{6.2981}} &  {{5.1429}} \\
          \cmidrule{2-13}
          & DPL\cite{15chen2021source}   & 0.7085 & 0.8548 & \underline{0.8865} & \underline{0.8195} & 0.8173 &       & \underline{5.4746} & \underline{4.8977} & \underline{4.2597} & 9.9130 & \underline{6.1363} \\
          & CBMT\cite{47tang2023source} & 0.8210 & \underline{0.8958} & \textbf{0.8901} & 0.7488 & \underline{0.8389} &       & 6.0618 & 11.9345 & 4.5245 & 13.1809 & 8.9254 \\
          & FSM\cite{10yang2022FSM}*   & \underline{0.8490} & 0.6160 & 0.7790 & 0.6730 & 0.7293 &       & 10.3940 & 10.1650 & 7.7740 & 5.3290 & 8.4155 \\
          & FVP\cite{11wang2023fvp}*   & \textbf{0.8560} & 0.7190 & 0.7950 & 0.6400 & 0.7525 &       & 9.0120 & 9.0030 & 4.3740 & \underline{3.5200} & 6.4773 \\
          &  {{DDFP (ours)}} &  {{0.7471}} &  {{\textbf{0.9049}}} &  {{0.8844}} &  {{\textbf{0.8770}}} &  {{\textbf{0.8534}}} &       &  {{\textbf{4.0955}}} &  {{\textbf{3.3774}}} &  {{\textbf{2.8331}}} &  {{\textbf{3.3440}}} & {{ \textbf{3.4125}}} \\
    \bottomrule
    \end{tabular}%
  \label{tab:tab2}%
\end{table*}%

\begin{table}[htbp]
  \centering
  \caption{{{Quantitative segmentation results on the brain tumor dataset. The best results are highlighted in bold.}}}
  \setlength{\tabcolsep}{0.006\columnwidth}{
    \begin{tabular}{ccccccc}
    \toprule
    \multicolumn{7}{c}{\textbf{Brain tumor}} \\
    \midrule
    \multirow{2}[4]{*}{Backbone} & \multirow{2}[4]{*}{Method} & \multicolumn{2}{c}{Flair→T2} &       & \multicolumn{2}{c}{T2→Flair} \\
\cmidrule{3-4}\cmidrule{6-7}          &       & Dice  & ASD   &       & Dice  & ASD \\
    \midrule
    \multirow{4}[3]{*}{{{U-Net}}} & {{Supervised}} & {{0.7393}} & {{4.2963}} &       & {{0.8343}} & {{2.3592}} \\
    & {{W/o adaptation}} & {{0.4669}} & {{14.3694}} &       & {{0.6149}} & {{11.3531}} \\
    \cmidrule{2-7}          & {{ProContra\cite{14yu2023source}}} & {{0.5433}} & {{12.4041 }}&       & {{0.5921}} & {{7.4121 }}\\
& {{DDFP (ours)}} & {{\textbf{0.6156}}} & {{\textbf{10.6579}}} &       & {{\textbf{0.7041}}} & {{\textbf{4.7802}}} \\
\midrule
\multirow{5}[2]{*}{{{DeepLab}}} & {{Supervised}} & {{0.7515}} & {{4.3463}} &       & {{0.8391}} & {{1.7730}} \\
      & {{W/o adaptation}} & {{0.4937}} & {{11.5511}} &       & {{0.5649}} & {{8.2796}} \\
\cmidrule{2-7}          & {{DPL\cite{15chen2021source}}}   & {{0.4802}} & {{11.0240}} &       & {{\textbf{0.7316}}} & {{4.5411}} \\
& {{CBMT\cite{47tang2023source}}}  & {{0.5274}} & {{13.7038}} &       & {{0.6773}} & {{6.7916}} \\
      & {{DDFP (ours)}} & {{\textbf{0.5995}}} & {{\textbf{9.3908}}} &       & {{0.7137}} & {{\textbf{4.0944}}} \\
    \bottomrule
    \end{tabular}}%
  \label{tab:addlabel}%
\end{table}%

Given that the magnitude of different losses varies across different datasets and adaptation directions, particularly the BN statistic loss, which is sensitive to domain gaps and adaptation difficulties, we use the loss function weights, {{rescaling the ratio of $ \mathcal{L}_{BNS}$: $ \mathcal{L}_{pseu}$: $ \mathcal{L}_{ent}$ to around 1, 0.01, 0.1,}} based on the values computed at the model's $0^{th}$ iteration (before using ground truth labels). For U-Net backbone multiorgan abdominal CT to MRI, MRI to CT, cardiac dataset CT to MRI, and MRI to CT adaptation, $[w_{ent},w_{BNS},w_{pseu}]$ are set to [1, 1, 10], [0.1, 1, 10], [4, 0.1, 10] (given that pseudo-labeling loss is more significant, so 0.1 is given), and [1, 1, 10], respectively. {{Those under Deeplab backbone are set to [1, 1, 10], [0.02, 1, 10], [4, 0.1, 10], [1, 1, 10], respectively. Besides, for the brain tumor datasets Flair to T2 and T2 to Flair adaptation under U-Net backbone, the loss weights are set to [5, 10, 20], [5, 1, 10], and those under Deeplab backbone are set to [0.1, 1, 10], [2, 1, 10], respectively.}} All experiments are conducted on a single NVIDIA GPU 3090 with Pytorch 1.12.1.

\subsubsection{Baselines}
We compared our method with several state-of-the-art SFDA methods, including prompt-based frameworks such as FVP \cite{11wang2023fvp} and FSM \cite{10yang2022FSM}, as well as self-supervised model fine-tuning methods such as ProContra \cite{14yu2023source}, DPL \cite{15chen2021source}, CBMT \cite{47tang2023source}, and TT-SFUDA \cite{48ttsfuda}. The results for FVP \cite{11wang2023fvp} are taken directly from the original paper, as the code is not publicly available. An asterisk (*) indicates results from \cite{11wang2023fvp}, which used random data partition and the same evaluation metrics. ``Supervised'' refers to the fully supervised results on the target domain, while ``W/o adaptation'' represents the result of directly applying the source model to the target domain without any adaptation.

\subsection{Results on the abdominal dataset}
The quantitative results for the CT to MRI and the MRI to CT adaptation on the abdominal dataset are presented in Table \ref{tab:tab1}. Our method achieves an average Dice score of 0.8987 for the CT to MRI adaptation, marking a 20-percentage point improvement compared to the ``W/o adaptation'' baseline. We also achieve the best average results in the ASD metrics. In the MRI to CT adaptation, the ``W/o adaptation'' performance is significantly lower than in the CT to MRI direction, with an average Dice score of only 0.4766, indicating that adaptation is more challenging in this direction. Our method achieves an average Dice score of 0.7934, outperforming the current state-of-the-art methods. While different methods exhibit varying performance across different organs, our DFFP primarily reduces the domain gap at the image level and does not incorporate boundary-level supervision. As a result, our method does not consistently improve both the Dice and ASD metrics in all cases. This aligns with observations from other prompt learning-based methods FVP \cite{11wang2023fvp}. The segmentation visualization results are shown in Fig. \ref{fig4}. It can be observed that our method accurately predicts the overall organ morphology. However, there are some deficiencies in the delineation of boundaries, especially in regions such as the spleen, which may explain the lower ASD scores observed for our method.

\begin{table}[t]
  \centering
  \caption{{{Ablation study results of different loss components.}} The best results are highlighted in bold.}
  \setlength{\tabcolsep}{0.008\columnwidth}{
    \begin{tabular}{cccccccc}
    \toprule
    \multirow{2}[4]{*}{$\mathcal{L}_{BNS}$} & \multirow{2}[4]{*}{$\mathcal{L}_{pseu}$} & \multirow{2}[4]{*}{$\mathcal{L}_{ent}$} & \multicolumn{5}{c}{\textbf{Dice ↑}} \\
\cmidrule{4-8}          &       &       & \textbf{Liver} & \textbf{R.kidney} & \textbf{L.kidney} & \textbf{Spleen} & \textbf{Average} \\
    \midrule
     
    {{\textbf{\ding{52}}}} &       &       & {{0.8154}} & {{0.9188}} & {{0.9187}} & {{0.8054}} &{{0.8646}} \\
          &  {{\textbf{\ding{52}}}} &       & {{0.8603}} & {{\textbf{0.9327}}} & {{0.9069}} & {{0.6944}} & {{0.8486}} \\
          {{\textbf{\ding{52}}}} &  {{\textbf{\ding{52}}}} &       & {{0.8672}} & {{0.9181}} & {{0.9235}} & {{0.8202}} & {{0.8822}} \\
     {{\textbf{\ding{52}}}} &       &  {{\textbf{\ding{52}}}} & {{0.8923}} & {{0.9122}} & {{0.8960}} & {{0.8059}} & {{0.8766}} \\
          &  {{\textbf{\ding{52}}}} &  {{\textbf{\ding{52}}}} & {{0.8752}} & {{0.9046}} & {{0.9129}} & {{0.8030}} & {{0.8739}} \\
          {{\textbf{\ding{52}}}} &  {{\textbf{\ding{52}}}} &  {{\textbf{\ding{52}}}} & {{\textbf{0.9053}}} & {{0.9206}} & {{\textbf{0.9263}}} & {{\textbf{0.8426}}} & {{\textbf{0.8987}}}  \\
    \bottomrule
    \end{tabular}}%
  \label{tab:tab3}%
\end{table}%
\newcommand{\tabincell}[2]{\begin{tabular}{@{}#1@{}}#2\end{tabular}}  

\subsection{Results on the cardiac dataset}
The quantitative results for the CT to MRI and MRI to CT adaptation on the cardiac dataset are shown in Table \ref{tab:tab2}. The average Dice score for the ``W/o adaptation'' baseline in the CT to MRI adaptation is only 0.4082, indicating difficulty in this adaptation direction for this dataset. By applying our method, the average Dice score improves to 0.6876, surpassing the performance of current state-of-the-art SFDA methods. Additionally, our method achieves a significant reduction in the average ASD, reaching 8.1182 mm, which is the best result among all the compared methods.

\begin{figure*}[!t]
  \centering
  \includegraphics[scale=0.257]{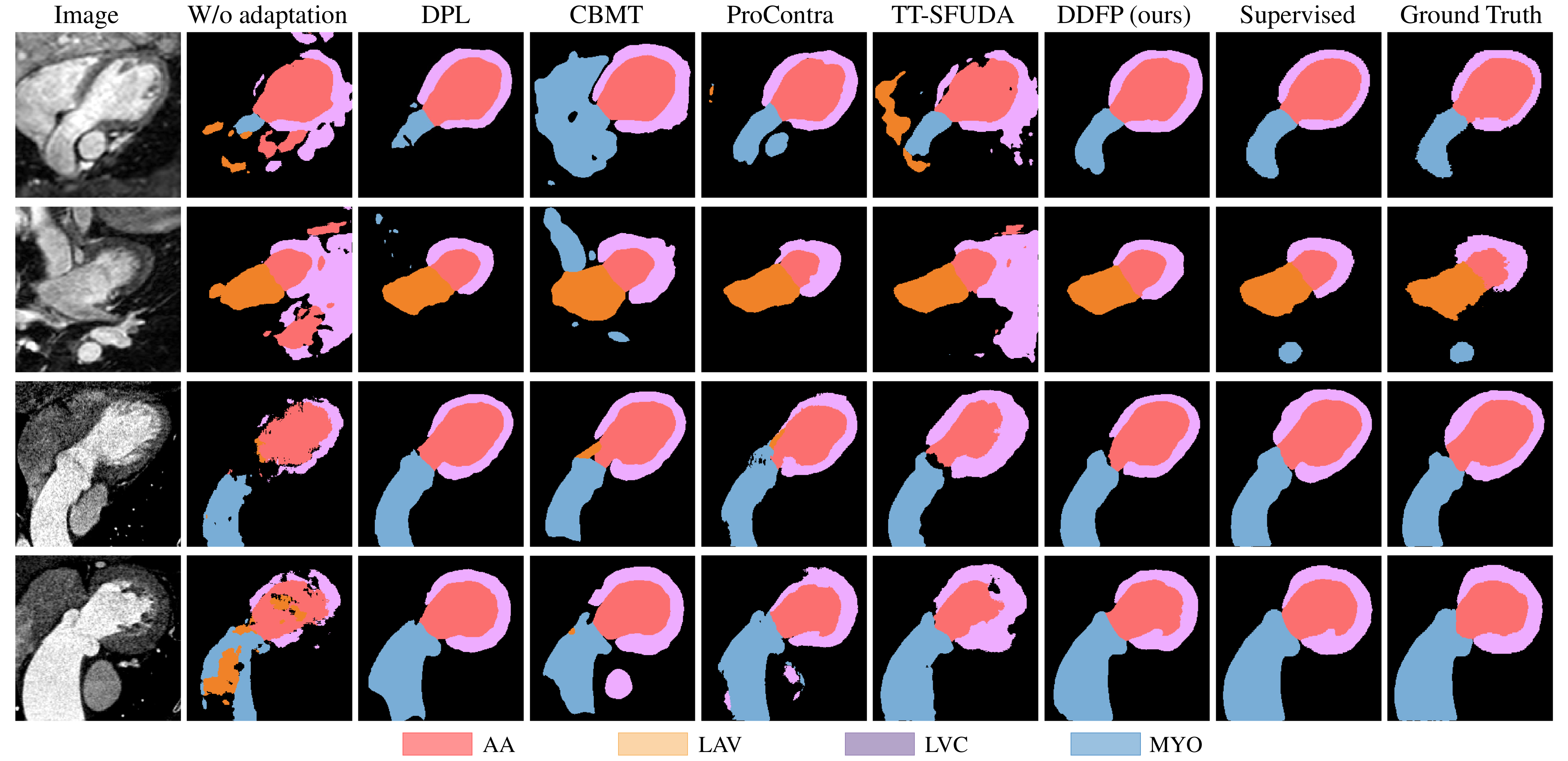}
  \caption{{{Visualization of SFDA segmentation results on the cardiac dataset.}} The first two rows correspond to CT to MRI adaptation, while the last two rows correspond to MRI to CT adaptation.}
  \label{fig5}
\end{figure*}

For the MRI to CT adaptation on the cardiac dataset, our method achieves an average Dice score of 0.8477 and an average ASD of 4.4150 mm, both of which are the best results among all the methods compared. The visualization results for the cardiac dataset are presented in Fig. \ref{fig5}. With the cardiac boundaries are relatively blurred and the segmentation task is more challenging compared to abdominal organ segmentation, our proposed method still provides a notable improvement in target model performance. It demonstrates a successful adaptation of the source model's knowledge to the target domain.

\subsection{Results on the brain tumor dataset}
The quantitative results for the T2 to Flair and Flair to T2 adaptation on the brain tumor dataset are shown in Table 3. The comparative method TT-FSUDA \cite{48ttsfuda} failed in the segmentation of some samples, therefore it is not included in the table. Our DDFP demonstrates better Dice and ASD results compared to the comparative methods in most cases.

\subsection{Statistic significant analysis}
\begin{figure}[!t]
  \centering
  \includegraphics[scale=0.138]{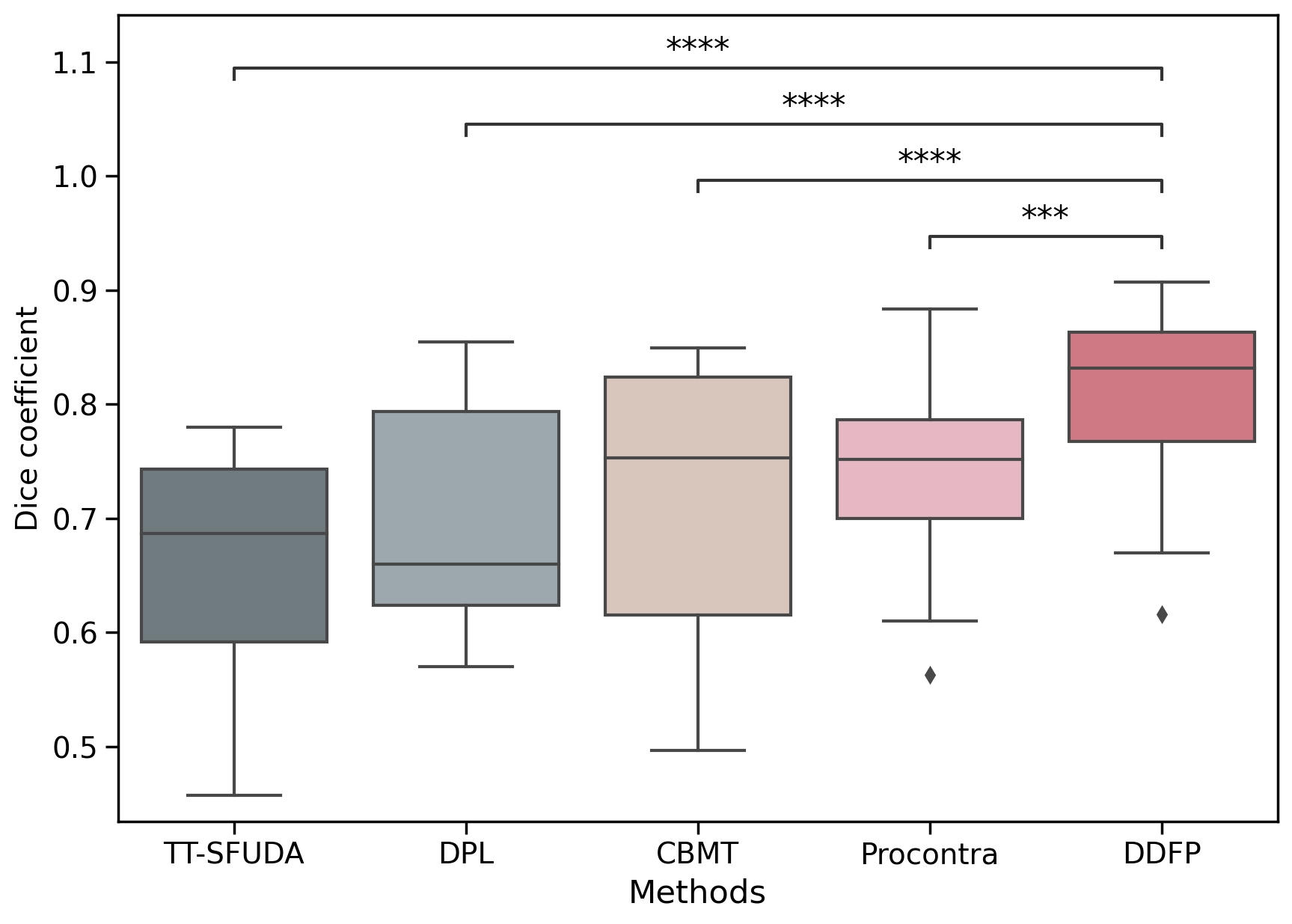}
  \caption{{{The boxplot results of experiments on the abdominal and cardiac datsets.} *: 1.00e-02 < p <= 5.00e-02, **: 1.00e-03 < p <= 1.00e-02,  **: 1.00e-04 < p <= 1.00e-03.}}
  \label{fig6}
\end{figure}

We performed a statistical significance test on the Dice coefficient. Given that our results are reported in 3D and the final number of test samples in each dataset and task is relatively small (around 5), this sample size is not ideal for statistical comparison. To provide a more comprehensive evaluation of our method's performance across different datasets and directions, we aggregated the results from both the cardiac and the abdominal datasets across all adaptation directions and conducted a statistical significance test on the 3D Dice coefficient. Due to the missing results of some methods on the brain tumor dataset, the restuls of BraTS2018 datasets are not included. Given that the data did not follow a normal distribution, we applied the Wilcoxon rank-sum test. The results, shown in Fig. \ref{fig6}, indicate that the differences between our method and the comparison methods are statistically significant, further validating the effectiveness of our approach.

\begin{table}[htbp]
  \centering
  \caption{Ablation study results of different prompting methods. ``Do.'' and ``Da.'' represent domain-dependent and data-dependent prompts, respectively. ``S.'' and ``F.'' refer to spatial and frequency domain prompts, respectively. The best results are highlighted in bold.}
  \setlength{\tabcolsep}{0.01\columnwidth}{
    \begin{tabular}{cccccc}
    \toprule
    \multirow{2}[4]{*}{\textbf{Prompt type}} & \multicolumn{5}{c}{\textbf{Dice ↑}} \\
\cmidrule{2-6}          & \textbf{Liver} & \textbf{R.kidney} & \textbf{L.kidney} & \textbf{Spleen} & \textbf{Average} \\
    \midrule
   Do., S. &  {{0.7607}} &  {{0.8406}} &  {{0.8836}} &  {{0.5478}} &  {{0.7582}} \\
    Do., F. &  {{0.7612}} &  {{0.8418}} &  {{0.8825}} &  {{0.5446}} &  {{0.7575}} \\
   Da., F. &  {{ \textbf{0.8008}}} &  {{\textbf{0.8771}}} &  {{\textbf{0.8994}}} &  {{\textbf{0.6585}}} &  {{\textbf{0.8090}}} \\
    \bottomrule
    \end{tabular}}%
  \label{tab:tab4}%
\end{table}%

\subsection{Ablation study}
\subsubsection{Loss components}
Table \ref{tab:tab3} presents the ablation results for the three components of the overall loss in the MRI to CT adaptation task on the abdominal dataset. The results show that using the BN layer statistic loss and pseudo-label loss separately already yields promising outcomes, with average Dice scores of 0.8646 and 0.8486, respectively. In contrast, using the entropy loss alone results in a Dice score of less than 0.3, likely due to the presence of semantic supervision, particularly in fine-tuning the style-dependent layers. As a result, the entropy loss alone is not included in the table. When combining two of the losses, the performance is comparable to or better than using each loss individually. The best performance, with an average Dice score of 0.8987, is achieved when all three losses are used simultaneously, providing comprehensive supervision across data distribution, feature semantics, and output entropy for both stylistic and semantic alignment.

\begin{table}[htbp]
  \centering
  \caption{Ablation study results of frequency prompt generation. The best results are highlighted in bold.}
  \setlength{\tabcolsep}{0.008\columnwidth}{
    \begin{tabular}{cccccc}
          \toprule
          \multirow{2}[4]{*}{\textbf{Operations}} & \multicolumn{5}{c}{\textbf{Dice $\uparrow$}} \\
      \cmidrule{2-6}          & \textbf{Liver} & \textbf{R.kidney} & \textbf{L.kidney} & \textbf{Spleen} & \textbf{Average} \\
      \midrule
       \textit{Components} &       &       &       &       &  \\
    W/o exp() & {{0.8789}} & {{0.9039}} & {{0.8839}} & {{0.7622}} & {{0.8572}} \\
    Only amplitude & {{0.8956}} & {{0.9001}} & {{0.9169}} & {{0.8025}} & {{0.8788}} \\
    \midrule
    {{\textit{Initalizations}}} &       &       &       &       &  \\
    {{ones}}  & {{0.8930}} & {{0.9013}} & {{0.8967}} & {{0.7506}} & {{0.8604}} \\
    {{rand}}  & {{0.8886}} & {{0.9139}} & {{0.9038}} & {{0.8141}} & {{0.8801}} \\
    {{Ours}}  & {{\textbf{0.9053}}} & {{\textbf{0.9206}}} & {{\textbf{0.9263}}} & {{\textbf{0.8426}}} & {{\textbf{0.8987}}} \\
    \bottomrule
    \end{tabular}}%
  \label{tab:tab5}%
\end{table}%

\subsubsection{Data-dependent frequency prompt}
We compared different prompting methods for the multiorgan abdominal CT to MRI adaptation task, including domain-dependent spatial prompts, domain-dependent frequency prompts, and the DFFP proposed in this work. Two sets of experiments were conducted: one with fine-tuning the style-related layers and one without. The results, shown in Table \ref{tab:tab4}, indicate that the DFFP achieves the best average Dice score. Given that frequency spectra are more closely related to the grayscale and style characteristics of images, frequency prompt learning provides better overall consistency in grayscale changes across images compared to spatial prompt learning. This aligns with previous findings \cite{11wang2023fvp}. Additionally, the data-dependent prompt effectively addresses internal variations within the dataset, leading to significant improvements in model performance. Visualization results are shown in Fig. \ref{fig8}.

Table \ref{tab:tab5} presents the ablation results of key components in the design of the DFFP. ``Only amplitude'' represents the scenario where only the amplitude spectrum is used to calculate the DFFP, as opposed to using both the amplitude and phase spectra. ``W/o $Exp()$'' denotes the case where the domain-dependent frequency prompt is applied directly without the exponential operator in Eq. \eqref{eq3}. The results demonstrate that the proposed DDFP framework achieves the best average Dice. This underscores score, highlighting the importance of jointly leveraging both spectra components. The main reason is that the amplitude and phase spectra mainly reflect. The amplitude spectrum primarily reflects grayscale information, while the phase spectrum encodes structural content. Together, these spectra capture intradomain variations more comprehensively. Therefore, using both spectra in the data-dependent prompt generation leads to more effective prompts, which in turn improves target model performance. Furthermore, the $Exp()$ operator constrains the prompted spectral values, facilitating more effective learning and optimization of the prompt.

{{
  Additionally, ablation experiments on abdominal CT to MRI adaptation under different initialization conditions are shown in Table \ref{tab:tab5}. Given that the computation of $FP^{data}$ in Eq. \eqref{eq4} involves the exponential of  $Exp(FP^{domain})$, initializing $FP^{domain}$ to all zeros results in $FP^{data}$ being initialized to nearly all ones. This initialization helps stabilize the outputs when multiplying $FP^{data}$ with the image's frequency spectrum.
}}

\begin{table*}[t!]
  \centering
  \caption{Ablation study results on the effect of BN preadaptation for target model initialization and pseudo-label generation. The best results are highlighted in bold.}
  \setlength{\tabcolsep}{0.07\columnwidth}{
    \begin{tabular}{ccccccc}
    \toprule
    \multirow{2}[4]{*}{{\tabincell{c}{\textbf{Target model}\\ \textbf{initialization}}}} & \multirow{2}[4]{*}{{\tabincell{c}{\textbf{Pseudo label}\\ \textbf{generation}}}} & \multicolumn{5}{c}{\textbf{Dice ↑}} \\
\cmidrule{3-7}          &       & \textbf{Liver} & \textbf{R.kidney} & \textbf{L.kidney} & \textbf{Spleen} & \textbf{Average} \\
    \midrule
    \multicolumn{2}{c}{W/o adaptation} &{{ 0.6198}} & {{0.3873}} & {{0.3541}} & {{0.5453}} & {{0.4766}} \\
    \midrule
  $\mathcal{M}_s$ & $\mathcal{M}_s$ & {{0.8714}}	& {{0.6762}}	& {{0.6265}}	& {{0.7610}}	& {{0.7338}}
  \\
  $\mathcal{M}_s^{'}$ & $\mathcal{M}_s$ & {{\textbf{0.8734}}}	& {{0.6841}}	& {{0.6349}}	& {{0.7569}}	& {{0.7373}}
  \\
  $\mathcal{M}_s^{'}$ & $\mathcal{M}_s^{'}$ & {{{0.8623}}} &	{{\textbf{0.7386}}} &	{{\textbf{0.7746}}} &	{{\textbf{0.7980}}}	& {{\textbf{0.7934}}}
  \\
    \bottomrule
    \end{tabular}}%
  \label{tab:tab6}%
\end{table*}%

\subsubsection{BN pre-adaptation}
In this experiment, we investigate the effect of the BN preadaptation strategy, which plays a crucial role in target model initialization and pseudo-label generation. We evaluate the impact of each role separately on the MRI to CT adaptation task using the abdominal dataset, with results presented in Table \ref{tab:tab6}. The results show that using the BN preadapted model for pseudo-label generation significantly outperforms using the source model, with the average Dice score improving from 0.7373 to 0.7934. Furthermore, initializing the target model with the BN preadapted model yields slightly better results than direct initializing the model with the source model. These findings indicate that the BN preadaptation strategy greatly enhances pseudo-label quality and the overall performance of the target model.

\begin{figure}[!t]
  \centering
  \includegraphics[scale=0.12]{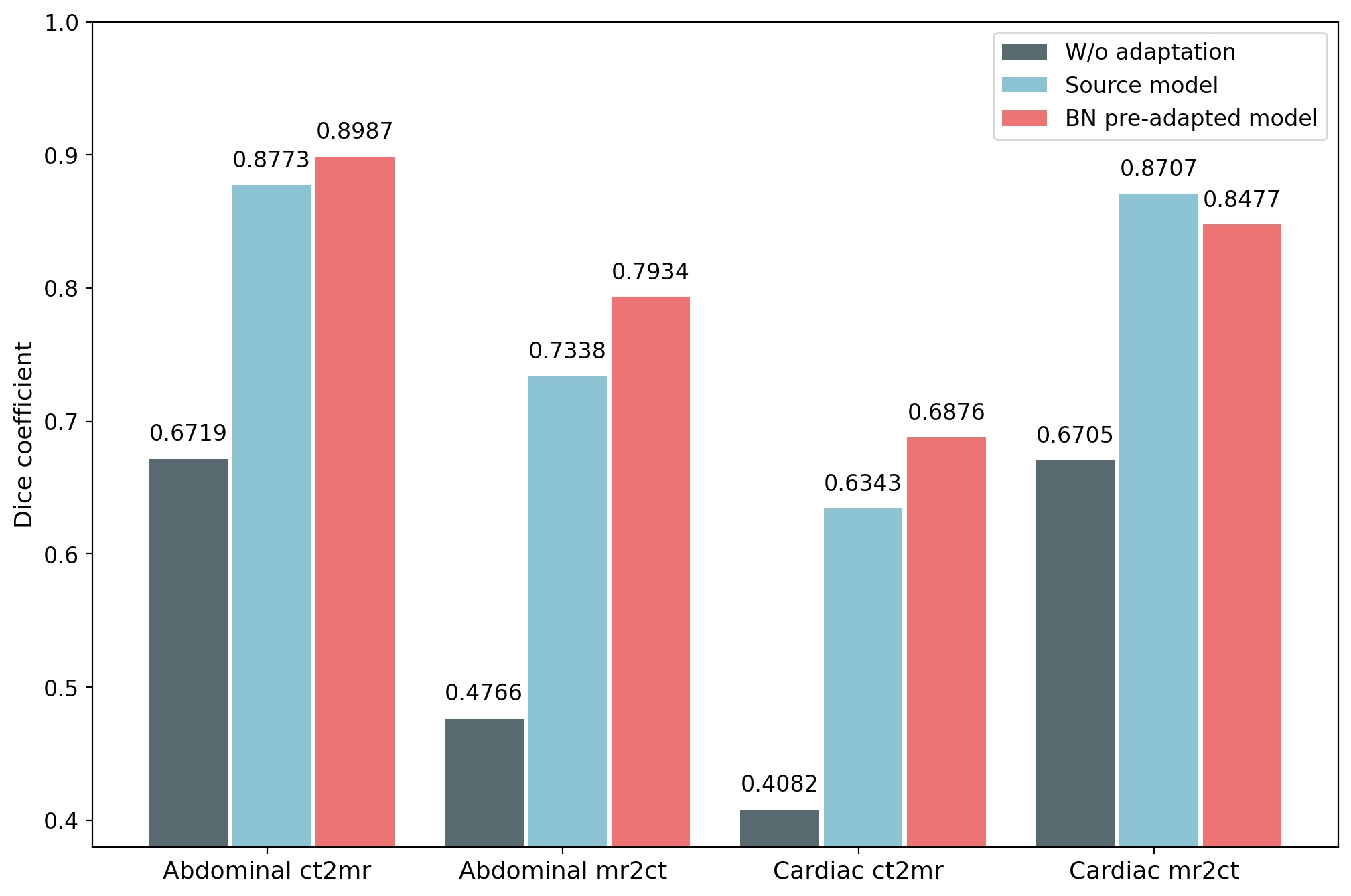}
  \caption{{{Results of different target model initialization approaches.}} Pink: Using the BN preadapted model for target model initialization and pseudo-label generation. Blue: Using the source model for target model initialization and pseudo-label generation. Gray: W/o adaptation.}
  \label{fig7}
\end{figure}

To assess how the effect of BN preadaptation might vary based on the difficulty of the adaptation task, we perform similar experiments in both adaptation directions across two datasets, with the results shown in Fig. \ref{fig7}. In the abdominal multiorgan CT to MRI adaptation and the cardiac MRI to CT adaptation, the improvement from BN preadaptation is modest, likely because these adaptations are relatively easier, as indicated by the higher performance of the ``W/o adaptation'' baseline. In contrast, in the more challenging adaptation tasks, such as the abdominal multiorgan MRI to CT and cardiac CT to MRI adaptations, the ``W/o adaptation'' performance drops significantly to 0.4766 and 0.4082, respectively. Under these more difficult adaptation conditions, using the BN preadaptation strategy for the target model initialization and pseudo-label generalization leads to a significant performance improvement.

\begin{figure}[!t]
  \centering
  \includegraphics[scale=0.32]{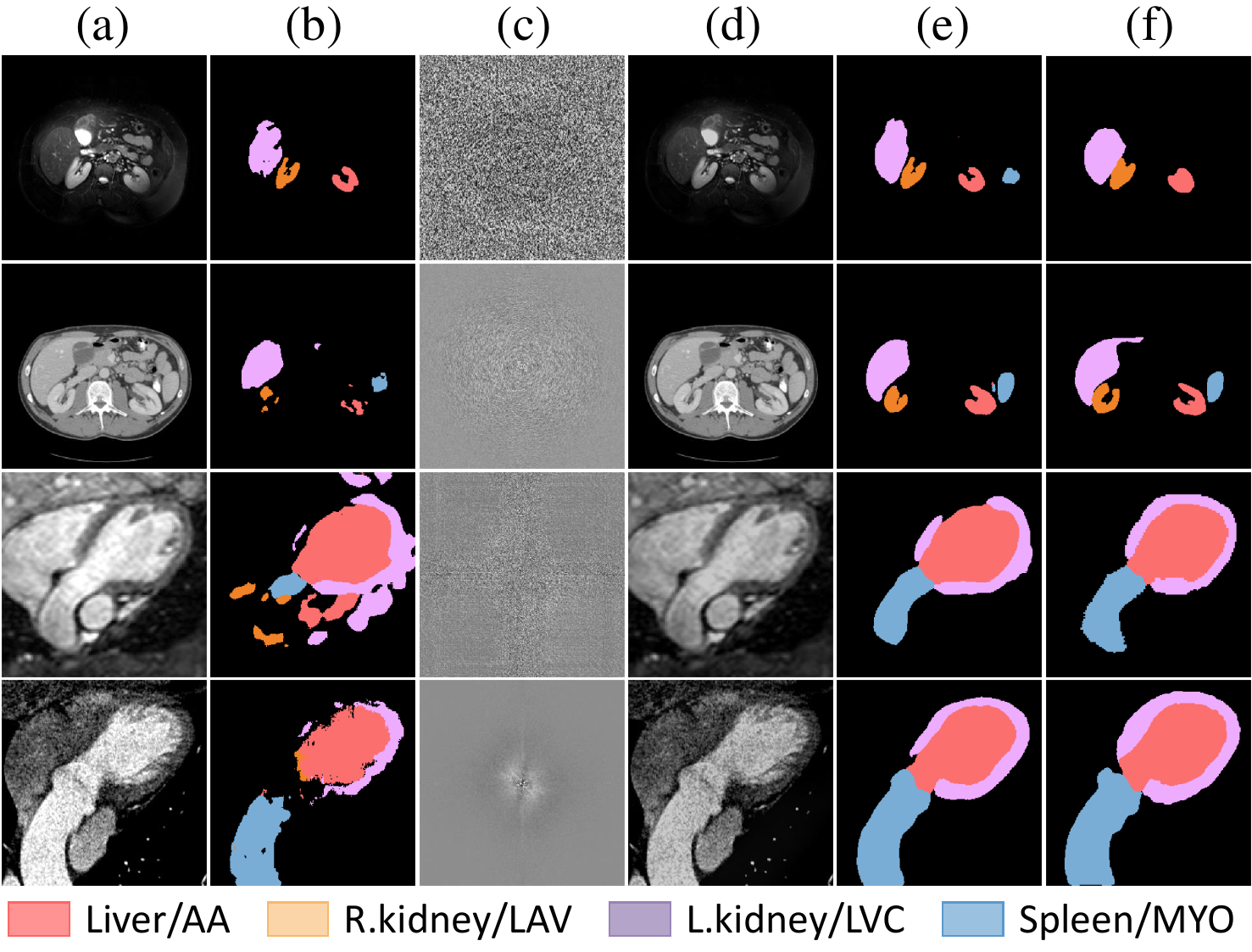}
  \caption{{{Visualization of data-dependent frequency prompts, pseudo-labels, and segmentation results.}} (a) Original image. (b) W/o adaptation result. (c) data-dependent frequency prompts. (d) Prompted image. (e) Segmentation results. (f) Ground truth.}
  \label{fig8}
\end{figure}

Finally, Figure  \ref{fig8} visualizes the DFFPs across diverse datasets and adaptation directions. The prompts primarily affect the low-frequency information region of the frequency spectrum, which is consistent with the fact that low-frequency information is closely tied to style characteristics in images.

\subsubsection{Setting of style-related layers}
To quantitatively evaluate the effectiveness of the style-related layer fine-tuning strategy in SFDA, we conduct experiments with different trainable layers in the target model, both with or without the DFFP, on the multiorgan abdominal CT to MRI adaptation task. The U-Net backbone consists of the $0^{th}$ convolutional layer (L0), three down-sampling convolutional layers (L1-3), and three up-sampling layers (L4-6), where layers L0-3 are considered style-related.

\begin{figure}[!t]
  \centering
  \includegraphics[scale=0.13]{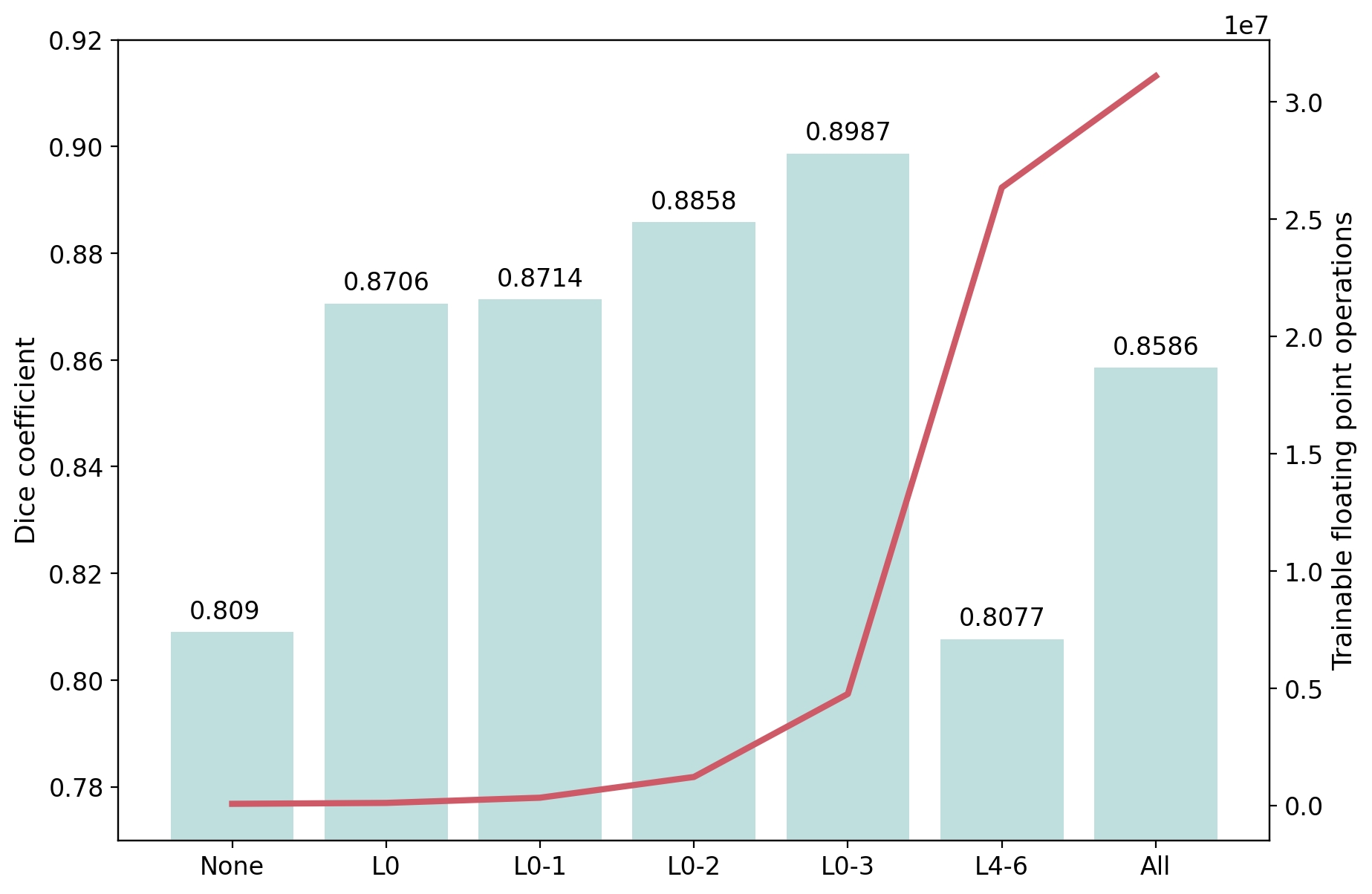}
  \caption{{{Trainable FLOPs for different trainable layers in the target model with the data-dependent frequency prompts.}}}
  \label{fig9}
\end{figure}

The results are shown in Fig. \ref{fig9}, along with the corresponding trainable floating-point operations (FLOPs). Training the style-related layers with the DFFP achieves an average Dice score of 0.8987, outperforming both training the entire model or other selection strategies. This demonstrates that the fine-tuning strategy used in our research not only achieves the best Dice score but also does so with a comparatively smaller number of parameters.

{\subsubsection{Hyperparameters}}
{{The parameter $\delta_{cls}$ is used to select the smallest $\delta_{cls}\%$ of pixels from each class, ensuring that reliable pseudo-labels are available for loss calculation in each class. This filtering prevents background pixels (which are abundant and easier to classify with lower entropy values) from dominating the reliable pseudo-labels. Experiment results using various $\delta_{cls}$ values (with $\delta_{glo}$ fixed at 0.4) and varying $\delta_{glo}$ values (with $\delta_{cls}$ fixed at 0.4) show that the choice of $\delta_{cls}$ has minimal impact on the results. More details and experiment findings are provided in the Supplementary material.}}

{{During the transfer training process, $\alpha$ is used in the calculation of $FP^{data}$ data to balance its contribution with the $FP^{domain}$ from the skip connection. $\theta$ is a scaling factor applied to $\mathcal{L}_{pseu}$. Experiments with different values for these two hyperparameters show that their specific choices have minimal impact on the results, with our approach consistently achieving superior performance regardless of the variations.}}

{{As for the loss weights used in Equ.\ref{eq14}, ablation results on abdominal CT to MR adaptation using [0.2, 0.5, 1, 5, 10] as the weight values (for w1, since the value is large, we used only 0.2, 0.5, and 1) are provided in the Supplementary material. The experiments demonstrate that adjusting the weights within a reasonable range does not significantly impact the results or conclusions. Since the weight design process here does not rely on true labels, when facing a new dataset, the same strategy can be used to set the weights, or the current settings can be applied, as they have little impact on the results.}}

\section{Discussion}
\textbf{Importance of this work.} Traditional unsupervised DA typically relies on labeled source domain data and unlabeled target domain data. However, in many real-world medical applications, privacy concerns can restrict access to source domain data. In such scenarios, SFDA becomes critical, as it enables DA using only the unlabeled target domain data and a pretrained source model. This makes SFDA more challenging but also highly relevant for practical applications.

\textbf{Benefits of key components.} We propose a novel framework for SFDA with three main contributions. First, we propose a DFFP, which effectively reduces the domain gap at the image level, outperforming previous domain-dependent prompting methods. Second, we introduce a BN preadaptation strategy that minimizes the domain gap early in the adaptation process. This improves pseudo-label quality and enhances target model performance without requiring additional training parameters, making it especially useful for large domain gaps. Third, we apply a style-related fine-tuning strategy tailored for SFDA, which optimizes model performance while minimizing the number of trainable parameters. Experiments on multiorgan abdominal and cardiac datasets validate the effectiveness of our approach.

\textbf{Limitation and future works.} Despite the promising results, some limitations remain. For example, the DFFP is currently fixed to the size of the input image, but exploring optimization of prompt size could further improve performance. Additionally, although the method improves Dice scores, some segmentation edges remain blurred, as the model lacks explicit edge constraints. {{Furthermore, applying our approach to classification tasks and extending it to other datasets could provide valuable insights and broaden its applicability.}}

\textbf{Conclusion.} This work introduces DDFP, a novel method for SFDA in medical image segmentation. We propose model preadaptation for target model initialization and pseudo-label generation, which enhances self-training performance by improving pseudo-label quality. Additionally, we introduce a DFFP for more effective image style translation and a style-related layer fine-tuning strategy for efficient target model training. Experimental results on multiorgan abdominal and cardiac SFDA tasks demonstrate the efficacy of our approach.

\printcredits

\bibliographystyle{cas-model2-names}

\bibliography{cas-refs}

\end{document}